\documentclass{article}

\PassOptionsToPackage{numbers, compress}{natbib}
\usepackage[preprint]{neurips_2024}

\usepackage[utf8]{inputenc} 
\usepackage[T1]{fontenc}    
\usepackage{hyperref}       
\usepackage{url}            
\usepackage{booktabs}       
\usepackage{amsfonts}       
\usepackage{nicefrac}       
\usepackage{microtype}      
\usepackage{xcolor}         
\usepackage{graphicx}
\usepackage{amsmath}
\usepackage{caption}
\usepackage[mathscr]{eucal}
\usepackage{multirow}
\usepackage{booktabs}
\usepackage{authblk}

\def\eg{\emph{e.g.}}
\def\ie{\emph{i.e.}}

\title{DA-Ada: Learning Domain-Aware Adapter for Domain Adaptive Object Detection}

\author{%
    \textbf{Haochen~Li\textsuperscript{1, 4}\quad Rui~Zhang\textsuperscript{2}\thanks{Corresponding author.}\quad Hantao~Yao\textsuperscript{3}\quad
    Xin~Zhang\textsuperscript{2}\quad
    Yifan~Hao\textsuperscript{2}\quad\vspace{-8pt}\\} \textbf{Xinkai~Song\textsuperscript{2}\quad  
    Xiaqing~Li\textsuperscript{2}\quad 
    Yongwei~Zhao\textsuperscript{2}\quad 
    Ling~Li\textsuperscript{1, 4}\footnote[1]{Corresponding author.}\quad Yunji~Chen\textsuperscript{2, 4}}
\\
    \textsuperscript{1}Intelligent Software Research Center, Institute of Software, CAS, Beijing, China\\
    \textsuperscript{2}State Key Lab of Processors, Institute of Computing Technology, CAS, Beijing, China\\
    \textsuperscript{3}
State Key Laboratory of Multimodal Artificial Intelligence Systems, Institute of Automation, CAS, Beijing, China\\
    \textsuperscript{4} University of Chinese Academy of Sciences, Beijing, China\\
    {\tt\small  haochen2021@iscas.ac.cn, zhangrui@ict.ac.cn, haotao.yao@nlpr.ia.ac.cn, }
    {\tt \small \{zhangxin, haoyifan, songxinkai, lixiaqing, zhaoyongwei\}@ict.ac.cn, liling@iscas.ac.cn, cyj@ict.ac.cn }  \\
}
\usepackage{tabularx}

\begin{document}

\maketitle

\begin{abstract}
Domain adaptive object detection (DAOD) aims to generalize detectors trained on an annotated source domain to an unlabelled target domain.
As the visual-language models (VLMs) can provide essential general knowledge on unseen images, freezing the visual encoder and inserting a domain-agnostic adapter can learn domain-invariant knowledge for DAOD.
However, the domain-agnostic adapter is inevitably biased to the source domain.
It discards some beneficial knowledge discriminative on the unlabelled domain, \ie domain-specific knowledge of the target domain.
To solve the issue, we propose a novel Domain-Aware Adapter (DA-Ada) tailored for the DAOD task.
The key point is exploiting domain-specific knowledge between the essential general knowledge and domain-invariant knowledge.
DA-Ada consists of the Domain-Invariant Adapter (DIA) for learning domain-invariant knowledge and the Domain-Specific Adapter (DSA) for injecting the domain-specific knowledge from the information discarded by the visual encoder.
Comprehensive experiments over multiple DAOD tasks show that DA-Ada can efficiently infer a domain-aware visual encoder for boosting domain adaptive object detection.
Our code is available at https://github.com/Therock90421/DA-Ada

\end{abstract}

\section{Introduction}
\label{sec:intro}
\begin{figure}[t]
\centering
\includegraphics[width=1.0\columnwidth]{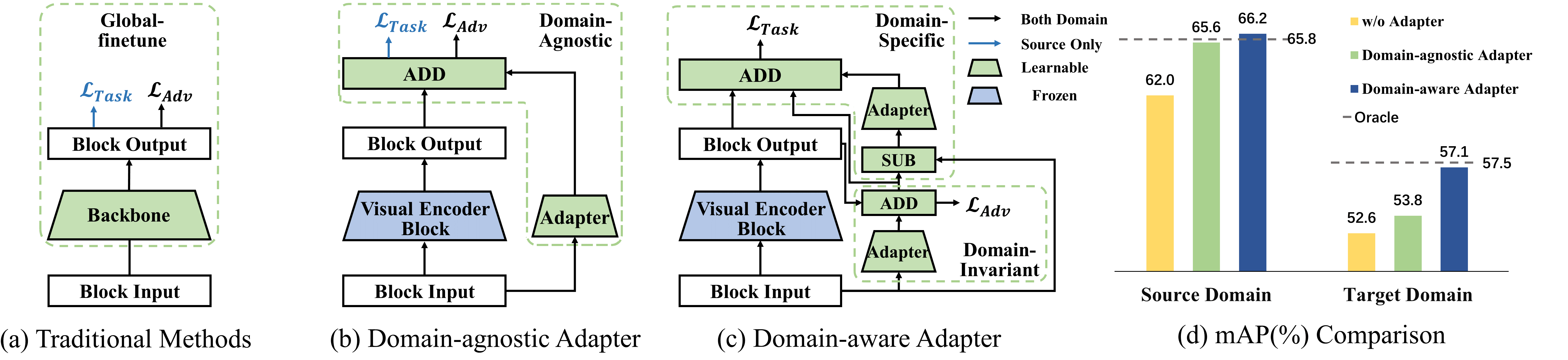}
\caption{(a) Traditional DAOD methods optimize the backbone adversarially. (b) Domain-agnostic adapter is inserted into the frozen visual encoder to learn domain-invariant knowledge. (c) Domain-aware adapter can simultaneously capture the domain-specific knowledge from the discarded feature. (d) The mAP($\%$) comparison on the Cross-Weather Adaptation. Compared with original VLM, domain-agnostic adapter brings significant improvement to the source domain but limited improvement to the source domain, while domain-aware adapter brings significant improvement to both source domain and target domain.}

\label{fig1}
\vspace{-25pt}
\end{figure}

Object detection~\cite{FasterRCNN, YOLO, FPN, ViT} have achieved remarkable performance, but suffer severe performance drop when dealing with unseen data due to domain discrepancy.
To alleviate this problem, domain adaptive object detection (DAOD)~\cite{DA-Faster} is explored to transfer a detector trained on the labelled source domain to a unlabelled target domain.
Traditional DAOD works~\cite{DA-Faster,VD,SAD,DSS,TIA,SIGMA,GPA,Category-contrast} generate the domain-aligned feature via fine-tuning the backbone, as depicted in Fig.~\ref{fig1}(a). 
Nevertheless, it is easily biased towards the source domain since a considerable number of parameters need to be updated with the annotations only from the source domain.

Recently, applying prompt tuning~\cite{CoOp,CoCoOp,KgCoOp} on visual-language models (VLMs) is widely used for two reasons: 1) few parameters need to be learned; 2) VLMs trained on large-scale image-text pairs can extract highly generalized features.
Recent works~\cite{DA-Pro, AD-Clip} have explored using prompt tuning to generate the domain-aware detection head for DAOD.
However, they all extract the visual feature from the image with a frozen visual encoder, ignoring learning task-related knowledge and limiting the improvement of visual features' discriminative capabilities.

To inject the task-related knowledge into the visual encoder, some methods~\cite{Adapter, V-adapter, Conv-adapter} insert an adapter module into the frozen backbone.
Formally, a adapter shared across domains can be straightly introduced to learn the task-related knowledge with the annotations of source domain and domain-aligned constraint, as shown in Fig.~\ref{fig1}(b).
However, this adapter is domain-agnostic and can only learn domain-invariant knowledge between the two domains under the domain-aligned constraint.
Besides, the domain-invariant knowledge is inevitably biased to the source domain, since the annotations are only from the source domain. 
As shown in Fig.~\ref{fig1}(d), compared with the original VLM, the domain-agnostic adapter brings significant improvement to the source domain, while the improvement of the target domain is limited. 
Summarily, the bias of the domain-invariant knowledge learned from domain-agnostic adapter limits the generalization to the unseen target domain. 

Trained on large-scale data, the VLM provides essential general knowledge on unseen images, while the learned domain-invariant knowledge biased to the source domain shows limited improvement on the target domain.
Consequently, when transferring essential general knowledge to the domain-invariant knowledge, the domain-agnostic adapter discards some beneficial knowledge on the target domain.
Basically, it discards the domain-specific knowledge that distinguishes the target domain but is different from the domain-invariant knowledge.
In summary, adding a complementary adapter to capture the target-specific knowledge from the discarded knowledge between the essential general knowledge and domain-invariant knowledge is an effective way to boost the performance of the VLM in DAOD task.


In this paper, we propose a novel Domain-aware Adapter (DA-Ada) to facilitate the visual encoder learning the domain-specific knowledge along with the domain-invariant knowledge.
Formally, DA-Ada introduces a Domain-Invariant Adapter (DIA) and Domain-Specific Adapter (DSA) to exploit domain-invariant and domain-specific knowledge, respectively, as shown in Fig.~\ref{fig1}(c).
The DIA is attached to the block of the visual encoder in parallel and optimized by aligning the feature distribution of two domains to learn domain-invariant knowledge.
The DSA is fed with the difference between the input and output of the block to recover the domain-specific knowledge discarded by the DIA.
Since the difference represents the feature discarded by the block, the discarded knowledge between the essential general knowledge and domain-invariant knowledge is also hidden in the difference.
Hence, the DSA can regain the domain-specific knowledge from the difference adaptively to improve the generalization ability on target domain, as shown in Fig.~\ref{fig1}(d).
Moreover, we propose the Visual-guided Textual Adapter (VTA), embedding cross-domain information learned by DA-Ada into textual encoder to enhance the discriminability of detection head.
Overall, the proposed DA-Ada can inject domain-invariant and domain-specific knowledge into VLM for DAOD.

We conduct evaluations on mainstream DAOD benchmarks: Cross-Weather (Cityscapes $\rightarrow$ Foggy Cityscapes), Cross-Fov (KITTI $\rightarrow$ Cityscapes), Sim-to-Real (SIM10K $\rightarrow$ Cityscapes) and Cross-Style (Pascal VOC $\rightarrow$ Clipart).
Experimental results show that the proposed DA-Ada brings noticeable improvement and outperforms state-of-the-art methods by a large margin.
For example, DA-Ada reaches $58.5\%$ mAP on Cross-Weather, surpassing the state-of-the-art DA-Pro~\cite{DA-Pro} by $2.7\%$.

\section{Related Work}
\label{sec:related}

\paragraph{Visual-Languague models}
Visual-language models (VLMs)~\cite{CLIP, BLIP, BLIP2} embed visual and text modalities into a shared space, enabling cross-modal alignment.
Pre-trained with an astonishing scale of image-text pairs, they demonstrate comprehensive visual understanding.
CLIP~\cite{CLIP} simultaneously trains a visual encoder and a textual encoder with 400 million image-text pairs, showing promising performance on both the seen and unseen classes.
Furthermore, \cite{Regionclip, ViLD, poda, ClipTheGap} distill the knowledge from the visual encoder of CLIP into the detection backbone and transform the textual encoder into detection head.
Considering strong generalization, we apply RegionCLIP~\cite{Regionclip} as the detector.

\paragraph{Domain Adaptive Object Detection (DAOD)} aims to adapt the object detector~\cite{FasterRCNN} trained on the labelled source domain to the unlabelled target domain.
Previous approaches can be broadly divided into two orthogonal categories: feature alignment and semi-supervised learning. 
Feature alignment~\cite{DAN,DTN,RTN,WMMD,FCP,DANN,JAN,SAP,DTN} aims to align the feature distributions of the two domains with domain discriminators~\cite{DA-Faster}, to generate domain-invariant knowledge in three levels: image-level~\cite{DA-Faster, VD, DSS, DIDN}, instance-level~\cite{RPN, Strong-weak} and category-level~\cite{MEGA-CDA, Category-contrast, sigma++}.
To prevent knowledge unique to each domain from interfering with alignment, recent works~\cite{DS, DSB, DSN, DSS, CDSD} propose multiple extractors\cite{DDF, DIDN, TFD, IIPD} and discriminators~\cite{TRKP} to decouple the domain-invariant and domain-specific knowledge.
In parallel, semi-supervised learning strives to augment training data with style transfer~\cite{FSAC, TDD, UMT} and pseudo label~\cite{MT, AT, PT, HT}.
However, applying existing DAOD method to VLM would overfit the model to the training data, compromising the generalization of pre-trained models.
To preserve the pre-trained knowledge, we opt to freeze the VLM and devise a novel domain-aware adapter to facilitate cross-domain adaptation.
Compared with existing decoupling methods that only use domain-invariant features for detection, our method adopts a decoupling-refusion strategy.
It adaptively modify domain-invariant features with domain-specific features to enhance the discriminability on the target domain.

\paragraph{Tuning method for VLM}
Adapting pre-trained VLM to downstream tasks via global finetuning is prohibitively expensive and easily overfitted to training datasets.
To solve this issue, prompt tuning~\cite{CoOp} replaces the hand-crafted prompts with the learnable tokens for the textual encoder.
Conditions like categories~\cite{CoCoOp}, human prior~\cite{KgCoOp} and domain knowledge~\cite{DA-Pro} are attached to attain robust performance on new tasks.
However, they freeze the visual encoder, preventing it from learning cross-domain information for DAOD.
In parallel, originated from Natural Language Processing (NLP)\cite{Adapter, adapterfusion, k-adapter,Parallel-adapter, unified}, adapter tuning inserts learnable small layers into the visual encoder so that the backbone can learn knowledge from new tasks.
ViT-Adapter~\cite{V-adapter} and Conv-Adapter~\cite{Conv-adapter} are proposed to efficiently transfer pre-trained knowledge to zero or few-shot visual tasks.
\cite{CLIP-Adapter} integrates the adapter into the CLIP model, and~\cite{VL-adapter} further analyzes the components to be frozen or learnable. 
\cite{SVL-Adapter} combines self-supervised learning to enhance the ability to extract low-level features. 
Recent~\cite{Medical-sam-adapter} explore injecting task-related knowledge into segmentation model SAM~\cite{SAM}. 
However, tuning the adapter directly on both domains will bias it towards the source domain and fails to distinguish domain-specific knowledge, leading to insufficient discrimination on the target domain.
In this paper, we propose a novel domain-aware adapter that explicitly learns both domain-invariant and domain-specific knowledge to inject cross-domain information into the visual encoder. 

\section{Methodology}
\label{sec:method}

In this section, we present a novel Domain-aware Adapter (DA-Ada) tailored for DAOD.
DA-Ada employs adapter tuning to introduce both domain-specific and domain-invariant knowledge into VLM.
It is worth noting that the proposed method can be attached to any CNN-based detectors as a plug-and-play module.
Without loss of generality, we take vanilla Faster-RCNN~\cite{FasterRCNN} as an example.
\vspace{-3pt}

\subsection{Overview}
\label{Overview}
Inspired by adapter tuning, we can custom learnable adapters to inject cross-domain information into the visual encoder.
Specifically, to enrich the extracted features with high domain generalization capabilities, an ideal adapter should satisfy conditions from the following two aspects.
First, it can model the commonalities between the source and target domains, \ie domain-invariant knowledge.
Second, it can adaptively supply the unique attributes of each domain, \ie domain-specific knowledge.

\begin{figure*}[t]
\centering
\includegraphics[width=1.0\textwidth]{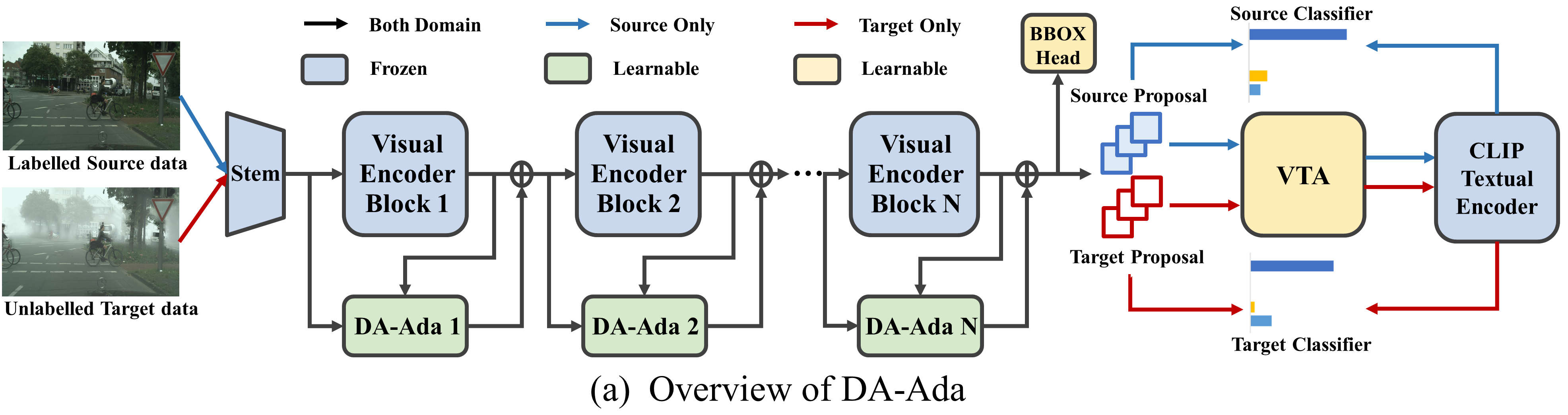} 
\vspace{3pt}
\centering
\includegraphics[width=0.9\columnwidth]{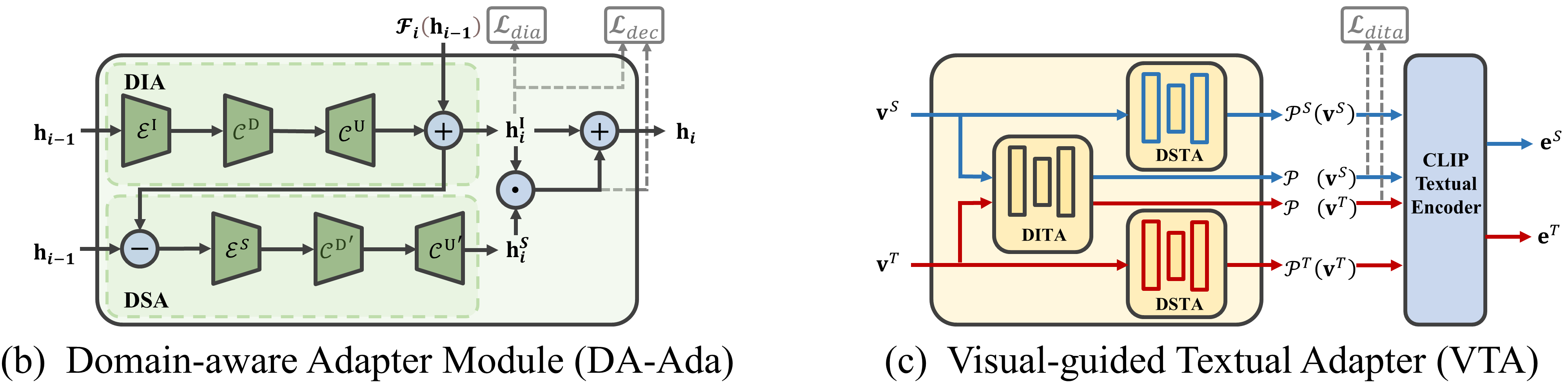} 
\vspace{-5pt}
\caption{Overview of the proposed (a) DA-Ada for DAOD and the architecture of (b) the $i$-th domain-aware adapter module (c) the visual-guided textual adapter.}
\label{fig-da-ada}
\vspace{-16pt}
\end{figure*}

In this perspective, we design an effective Domain-Aware Adapter (DA-Ada) consisting of a Domain-Invariant Adapter (DIA) and a Domain-Specific Adapter (DSA).
As shown in Fig.~\ref{fig-da-ada}(a), given input image $\mathbf{x}$, we split the visual encoder into $N$ blocks $\{\mathscr{F}_i\}_{i=1}^N$ by feature resolutions ($N$ = 4 in ResNet).
Then we attach $N$ blocks with DA-Ada modules $\{\mathscr{A}_i\}_{i=1}^N$ in Fig.~\ref{fig-da-ada}(b):
\begin{equation}
    \mathbf{h}_0 = \mathscr{S}(\mathbf{x}); \mathbf{h}_i = \mathscr{A}_i(\mathbf{h}_{i-1}, \mathscr{F}_i(\mathbf{h}_{i-1})),
\end{equation}
where $\mathscr{S}$ denotes the stem layer.
For the $i$-th DA-Ada module, we first feed the $i$-th block's input $\mathbf{h}_{i-1}$ into the $i$-th DIA module $\mathscr{A}_i^I$ to extract the domain-invariant features $\mathbf{h}^I_i$.
Then we attain the domain-specific features $\mathbf{h}^S_i$ from the subtraction of $\mathbf{h}_{i-1}$ and $\mathbf{h}^I_i$ by the DSA module $\mathscr{A}_i^S$:
\begin{equation}
    \label{Inv}
    \mathbf{h}_i^{I} = \mathscr{A}_i^I(\mathbf{h}_{i-1}) + \mathscr{F}_i(\mathbf{h}_{i-1});\\
    \mathbf{h}_i^{S} = \mathscr{A}_i^S(\mathbf{h}_{i-1} - \mathbf{h}_i^I).
\end{equation}
\vspace{-10pt}

After that, we fuse  $\mathbf{h}_i^I, \mathbf{h}_i^S$ with spatial attention to output $\mathbf{h}_i$ for $i$-th block: 
\begin{equation}
    \label{da-ada}
    \mathbf{h}_i = \mathbf{h}_i^I + \mathbf{h}_i^I\cdot \mathbf{h}_i^S,
\end{equation}
where $\cdot$ denotes the element-wise Hadamard product.
With $N$ learnable adapters, we obtain visual embedding $\mathbf{v}=\mathbf{h}_N$ for subsequent detection.
As the visual embedding contains sufficient cross-domain information, we propose the Visual-guided Textual Adapter (VTA), projecting the visual embedding to the textual encoder to enhance the discriminability of the detection head.
As shown in Fig.~\ref{fig-da-ada}(c), the visual-guided textual adapter uses the visual embedding $\mathbf{v}^S, \mathbf{v}^T$ to infer textual embedding $\mathbf{e}^S, \mathbf{e}^T$ on source and target domain, which is utilized for prediction.
Overall, the proposed DA-Ada can inject domain-invariant and domain-specific knowledge into VLM to improve cross-domain generalization ability.

\vspace{-5pt}
\subsection{Domain-Invariant Adapter (DIA)}
The DIA module is proposed to inject the domain-invariant knowledge into the visual encoder.
As shown in Fig.~\ref{fig-da-ada}(b), it applies a bottleneck to learn multi-scale domain knowledge, and the output distribution is aligned between domains for extracting domain-invariant knowledge.
Specifically, for the $i$-th block of the visual encoder, we first forward the input feature $\mathbf{h}_{i-1} \in \mathbb{R}^{b\times c\times h\times w}$ to the $i$-th DIA and filter domain-irrelevant information with embedding block $\mathscr{E}^{I}$:
\begin{equation}
    \mathbf{h}_i^E = \mathscr{E}^I(\mathbf{h}_{i-1}).
\end{equation}

After that, the embedding $\mathbf{h}_i^E \in \mathbb{R}^{b\times c\times h\times w}$ is helpful for domain representation learning.
Low channel-dimensional features have less information redundancy and are more suitable for domain adaptation than high-dimensional ones.
Following this spirit, the embedding is encouraged to be down-projected to a low channel-dimensional vector $\mathbf{h}_i^L\in \mathbb{R}^{b\times r\times h\times w}$ to extract domain-invariant knowledge and filter redundant information.
Formally, a down-projection $\mathscr{C}^{D}$ is applied to reduce the dimension to $r$:
\begin{equation}
    \label{Hidden}
    \mathbf{h}_i^L = \mathscr{C}^{D}(\mathbf{h}_i^E).
\end{equation}

Considering that the scale of objects varies between domains, we introduce $M$ down-projectors $\{\mathscr{C}^{D}_i\}_{i=1}^M$ with different receptive fields, enabling it to capture various spatial features across multiple scales.
Specifically, the embedding ${\mathbf{h}_i^E}=[\mathbf{h}_{i,1}^{E}, \mathbf{h}_{i,2}^{E},..., \mathbf{h}_{i,M}^{E}]$ is first split up evenly in the channel dimension.
Then, each partition is resized to different resolutions and down-projected.
Therefore, the multi-scale version of Eq.~\eqref{Hidden} is expressed as:
\begin{equation}
    \mathbf{h}_i^L = [\mathscr{C}^{D}_1(\mathbf{h}_{i,1}^{E}), \mathscr{C}^{D}_2(\mathbf{h}_{i,2}^{E})..., \mathscr{C}^{D}_M(\mathbf{h}_{i,M}^{E})].
\end{equation}

Furthermore, the low-dimensional knowledge $\mathbf{h}_i^L$ is encouraged to be mapped back to the original dimensional feature space and supplemented to the pre-trained features.
Typically, we apply the dimension-raising function $\mathscr{C}^U$ on $\mathbf{h}_i^L$ to extract domain-invariant knowledge for the visual encoder.
\begin{equation}
    \mathscr{A}_i^I(\mathbf{h}_{i-1}) = \mathscr{C}^U(\mathbf{h}_i^L),
\end{equation}
where $\mathscr{A}_i^I(\mathbf{h}_{i-1})\in \mathbb{R}^{b\times c\times h\times w}$ is output of the $i$-th DIA, and will be summed with $\mathscr{F}_i^I(\mathbf{h}_{i-1})$ to attain domain-invariant feature $\mathbf{h}_{i}^I$ in Eq.~\eqref{Inv}.
To ensure DIA learning domain-invariant knowledge, the $\mathbf{h}_{i}^I$ is expected to be well aligned between the two domains.
Therefore, $N$ domain discriminator $\{\mathscr{D}_i\}_{i=1}^N$ is attached to each $\mathbf{h}_{i}^I$ to calculate adversarial loss $\mathscr{L}_{dia}$. 
We will introduce this loss in Sec.\ref{Optimization Objective}.

With the combination of dimensional reduction-increase processes and the constraints of detection and adversarial loss, the DIA can extract domain-invariant features while reducing redundant features.

\subsection{Domain-Specific Adapter (DSA)}
Adapted with DIA, the essential general knowledge of the frozen VLM is transferred to domain-invariant knowledge.
However, the knowledge learned only through the DIA is biased towards the source domain and appears less discriminative on the target domain.
Considering the high generalization of essential general knowledge of the frozen VLM, we attribute this problem to the fact that the DIA discards some domain-specific knowledge that is highly generalizable on the unlabelled target domain.
Since the difference between the input and output of the block denotes the discarded feature, the discarded domain-specific knowledge is also hidden in the difference.
To this end, we propose the DSA module to recover domain-specific knowledge from the difference.

After the DIA injects the domain-invariant knowledge into the visual encoder, the domain-specific knowledge unique to the target domain is discarded by the output $\mathbf{h}_{i}^I$.
Therefore, we first obtain the feature $\mathbf{h}_i^D$ discarded by the visual encoder block from the difference of the input $\mathbf{h}_{i-1}$ and $\mathbf{h}_{i}^I$:
\begin{equation}
    \mathbf{h}_i^D = \mathscr{E}^{S}(\mathbf{h}_{i-1} - \mathbf{h}_{i}^I), 
\end{equation}
where $\mathscr{E}^{S}$ is similar with the embedding block $\mathscr{E}^{I}$. 
As domain-specific knowledge is hidden in the discarded difference $\mathbf{h}_i^D$, a bottleneck architecture is employed for adaptive knowledge extraction:
\begin{equation}
    {\mathbf{h}^L}'_i = {\mathscr{C}^D}'(\mathbf{h}_i^D),
\end{equation}
\vspace{-10pt}
\begin{equation}
    \mathbf{h}_{i}^S = \mathscr{A}_i^S(\mathbf{h}_{i-1} - \mathbf{h}_i^I) = {\mathscr{C}^U}'({\mathbf{h}^L}'_i),
\end{equation}
where ${\mathscr{C}^D}', {\mathscr{C}^U}'$ follow the same configurations as ${\mathscr{C}^D}, {\mathscr{C}^U}$ to perceive multi-scale domain-specific knowledge in bottleneck manner.

Generally speaking, domain-invariant knowledge dominates the process of transferring essential general knowledge of the VLM, while domain-specific knowledge fine-tunes this process based on the characteristics of each domain.
To this end, it is a more reasonable way to adaptively supplement domain-specific knowledge with the extracted $\mathbf{h}_{i}^S$ through pixel-level attention rather than straightforward addition.
Therefore, the injection of the whole DA-Ada is written as Eq.~\eqref{da-ada}.

\vspace{-5pt}
\subsection{Visual-guided Textual Adapter (VTA)}
\label{visual-guided textual adapter}
\begin{figure*}[t]
\centering
\includegraphics[width=0.95\textwidth]{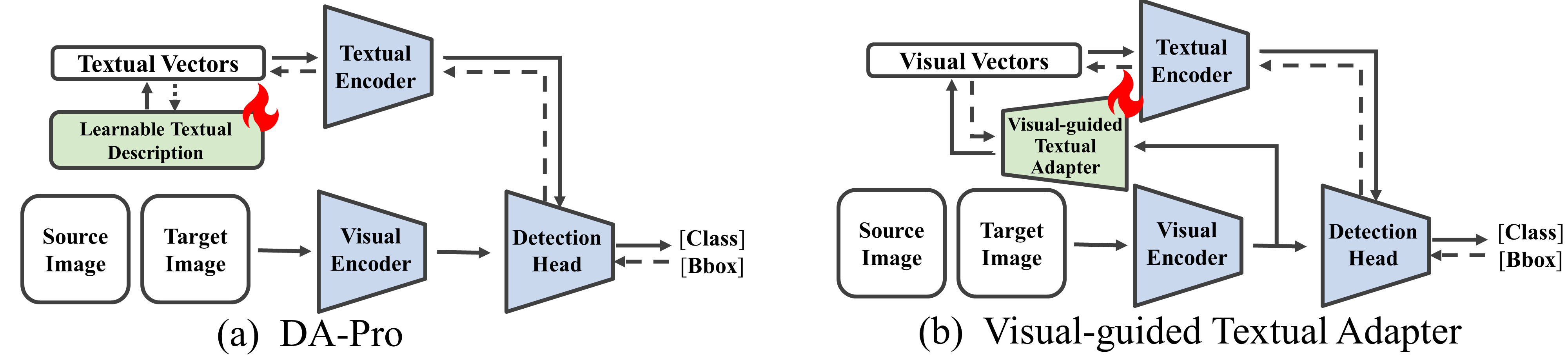} 
\caption{Comparison between (a) DA-Pro and (b) Visual-guided textual adapter.}
\label{fig-difference}
\vspace{-15pt}
\end{figure*}

With the domain-aware adapter to inject domain-invariant and domain-specific knowledge, the extracted visual feature shows rich discriminability, which can also be used to improve detection head. 
Therefore, we introduce the VTA to exploit the cross-domain information contained in the visual features to enhance the textual encoder.

In order to fully exploit the domain-invariant and domain-specific knowledge extracted by the DA-Ada module, we equip two learnable components for VTA: the domain-invariant textual adapter $\mathscr{P}$ (DITA) and the domain-specific textual adapter $\mathscr{P}^S, \mathscr{P}^T$ (DSTA) shown in Fig.~\ref{fig-da-ada}(c).
The DITA is shared across domains to encode visual domain-invariant knowledge into the input of the textual encoder, optimized by a domain discriminator $\mathscr{D}^P$.
The DSTA is tailored to further supplement domain-specific knowledge for the source domain $S$ and the target domain $T$.
In practice, the structure of DITA and DSTA is a 3-layer MLP with a hidden dimension of 512, projecting visual embeddings into 8 tokens for the textual encoder. 
Formally, the VTA embeds visual information into the textual embedding,
\begin{equation}
    \label{eq3}
        \mathbf{e}^{S}_i=\mathscr{T}(\mathscr{P}(\mathbf{v}^S),\mathscr{P}^S(\mathbf{v}^S),c_i);\\
        \mathbf{e}^{T}_i=\mathscr{T}(\mathscr{P}(\mathbf{v}^T),\mathscr{P}^T(\mathbf{v}^T),c_i),
\end{equation}
where $\mathbf{v}^S, \mathbf{v}^T$, $i$ and $c_i$ denote the visual embedding from domain $S,T$, the $i$-th class and its textual description.
$\mathscr{T}$ denotes the textual encoder.
$\mathbf{e}^{S}_i$ and $\mathbf{e}^{T}_i$ is the textual-level classifier embedding of $i$-th class from the source and target domains, respectively.

Our proposed VTA introduces discriminative visual features into the textual encoder, alleviating the problem of insufficient adaptation in plain textual tuning.
Existing methods~\cite{DA-Pro} only tune learnable textual descriptions for detection head, as shown in Fig.~\ref{fig-difference}(a).
However, textual descriptions are insufficient to describe certain inter-domain differences, \eg, differences in fields of view, leading to a limited ability to learn cross-domain information.
Different from them, VTA analyses domain-invariant and domain-specific knowledge from visual features, inferring an image-conditional detection head with high discriminability, as shown in Fig.~\ref{fig-difference}(b).

\vspace{-5pt}
\subsection{Optimization Objective}
\label{Optimization Objective}
We aim to insert the DA-Ada into the visual encoder to learn cross-domain information and further tune the prompt for discriminative textual representation with image conditions.
On the one hand, we introduce domain adversarial loss to the DIA and DITA to guide the learning of domain-invariant information. 
Formally, we obtain the output features $\mathbf{h}_i^{I,S}, \mathbf{h}_i^{I,T}$ for the source image $\mathbf{x}_s$ and the target image $\mathbf{x}_t$ of each DIA, and minimize the adversarial loss:
\vspace{-5pt}
\begin{equation}
    \mathscr{L}_{dia}=-\sum_{i=1}^N[\mathbb{E}_{\mathbf{x}_s}||\mathscr{D}_i(\mathbf{h}_i^{I,S})||^2_2+\mathbb{E}_{\mathbf{x}_t}||\mathscr{D}_i(\mathbf{h}_i^{I,T})-\mathbf{1}||^2_2].
\end{equation}
\vspace{-5pt}
And the domain-shared DITA is expected to be aligned between domains:
\begin{equation}
    \mathscr{L}_{dita}=-[\mathbb{E}_{\mathbf{x}_s}||\mathscr{D}^P(\mathbf{v}^S)||^2_2+\mathbb{E}_{\mathbf{x}_t}||\mathscr{D}^P(\mathbf{v}^T)-\mathbf{1}||^2_2],
\end{equation}
where $\mathbf{v}^S, \mathbf{v}^T$ denotes the source and target visual embedding.

On the other hand, we learn task-related domain-specific knowledge in a semi-supervised manner.
For the source image, we calculate the cross-entropy for each visual embedding $\mathbf{v}^S$ with its annotations $y$.
For the target $\mathbf{v}^T$, we first obtain the prediction via the hand-crafted prompt "A photo of [class]" and filter out high-confidence pseudo labels $y'$, then minimize the cross-entropy as well:
\begin{equation}
    \mathscr{L}_{det} = \mathscr{L}_{ce}(\mathbf{v}^S\times \mathbf{e}^S,y) + \mathscr{L}_{ce}(\mathbf{v}^T\times \mathbf{e}^T,y'),
\end{equation}
where $\times$ denotes Matrix multiplication.

Meanwhile, to decouple domain-invariant and domain-specific knowledge, we maximize the distribution discrepancy between DIA and DSA.
\vspace{-5pt}
\begin{equation}
    \mathscr{L}_{dec} = \sum_{i=1}^N[\mathbb{E}_{\mathbf{x}_s,\mathbf{x}_t}max[0,cos(\mathbf{h}_i^I, \mathbf{h}_i^I\cdot\mathbf{h}_i^S) - \beta]],
\end{equation}
where $cos(\mathbf{a},\mathbf{b})=\frac{|\mathbf{a}^\top\cdot\mathbf{b}|}{||\mathbf{a}||^2_2||\mathbf{b}||^2_2}$ is absolute value of cosine distance, $\beta$ is a threshold.

With the help of domain classifiers, DIA and DITA are encouraged to contain more domain-invariant knowledge.
By minimizing $\mathscr{L}_{dec}$, the gap between DIA and DSA will be enlarged, which promotes DSA to extract more domain-specific knowledge.
Overall, the optimization objective is:
\begin{equation}
    \mathscr{L} = \mathscr{L}_{det} + \lambda_{dia}\mathscr{L}_{dia} + \lambda_{dita}\mathscr{L}_{dita} + \lambda_{dec}\mathscr{L}_{dec} + \mathscr{L}_{reg},
\end{equation}
where $\mathscr{L}_{reg}$ is the regression loss, and $\lambda_{dia}, \lambda_{dita}, \lambda_{dec}$ are balance ratios.

\vspace{-2pt}

\section{Experiment}
\subsection{Datasets and Implementation}
We evaluate our method on four benchmarks: Cross-Weather(Cityscapes~\cite{Cityscapes}$\rightarrow$Foggy Cityscapes~\cite{FoggyCityscapes}), Cross-Fov(KITTI~\cite{KITTI}$\rightarrow$Cityscapes), Sim-to-Real(SIM10k~\cite{SIM10K}$\rightarrow$Cityscapes) and Cross-Style(Pascal VOC~\cite{pascal}$\rightarrow$Clipart~\cite{clipart}).
Following~\cite{DA-Pro}, we adapt RegionCLIP(ResNet-50~\cite{Resnet}) with Faster-RCNN architecture as the baseline detector.
We detail the datasets and implementation in Sec.~\ref{Dataset}and~\ref{SOTA} of the Appendix.
\vspace{-5pt}

\begin{table*}[t]
\centering
\caption{\small Comparison ($\%$) with existing methods on Cross-Weather Cityscapes$\rightarrow$Foggy Cityscapes (C$\rightarrow$F), Cross-Fov KITTI$\rightarrow$Cityscapes (K$\rightarrow$C) and Sim-to-Real adaptation SIM10K$\rightarrow$Cityscapes (S$\rightarrow$C). * denotes CLIP~\cite{CLIP}-based methods.}
\label{tab1}
\vspace{-3pt}
\setlength\tabcolsep{6pt}
\resizebox{0.95\textwidth}{!}{ 
\begin{tabular}{cccccccccccc}
\toprule
   &\multicolumn{9}{c}{C$\rightarrow$F} & K$\rightarrow$C & S$\rightarrow$ C  \\
    \cmidrule(r){2-10} \cmidrule(r){11-11} \cmidrule(r){12-12}
Methods&Person&Rider&Car&Truck&Bus&Train&Motor&Bicycle&mAP&mAP&mAP\\
\midrule
DA-Faster~\cite{DA-Faster} &29.2 &40.4 &43.4 &19.7 &38.3 &28.5 &23.7 &32.7 &32.0 &41.9 & 38.2\\
SIGMA++~\cite{sigma++} & 46.4 & 45.1 & 61.0 & 32.1 & 52.2 & 44.6 & 34.8 & 39.9 & 44.5 &49.5 & 57.7\\
CIGAR~\cite{CIGAR} &46.1 &47.3 &62.1 &27.8 &56.6 &44.3 &33.7 &41.3 &44.9 & 48.5& 58.5\\
CSDA~\cite{CSDA} &46.6 &46.3 &63.1 &28.1 &56.3 &53.7 &33.1 &39.1 &45.8& 48.6& 57.8\\
HT~\cite{HT} &52.1 &55.8 &67.5 &32.7 &55.9 &49.1 &40.1 &50.3 &50.4 & 60.3 & 65.5\\
D$^2$-UDA~\cite{D2-UDA} &46.9 &53.3 &64.5 &38.9 &61.0 &48.5 &42.6 &54.2 &50.6 & 60.3 & 58.1\\
AT~\cite{AT} &56.3 &51.9 &64.2 &38.5 &45.5 &55.1 &\textbf{54.3} &35.0 &50.9 &-&- \\
NSA-UDA~\cite{NSA}  &50.3 &60.1 &67.7 &37.4 &57.4 &46.9 &47.3 &54.3 &52.7& 55.6& 56.3\\
DA-Pro~\cite{DA-Pro}*  &55.4 &62.9 &70.9 &40.3 &63.4 &54.0 &42.3 &58.0 &55.9& 61.4& 62.9\\
\midrule
DA-Ada(Ours)*  &\textbf{57.8} &\textbf{65.1} &\textbf{71.3} &\textbf{43.1} &\textbf{64.0} &\textbf{58.6} &48.8 &\textbf{58.7} &\textbf{58.5}& \textbf{66.7}& \textbf{67.3}\\
\bottomrule
\end{tabular}  
}
\vspace{-6pt}
\end{table*}

\begin{table*}[t]
\centering
\caption{\small Comparison ($\%$) with existing methods on Cross-Style adaptation task Pascal VOC$\rightarrow$Clipart.  * denotes CLIP~\cite{CLIP}-based methods.}
\label{tab2}
\vspace{-2pt}
\setlength\tabcolsep{1.5pt}
\resizebox{0.95\textwidth}{!}{      
\begin{tabular}{cccccccccccccccccccccc}
\toprule
Methods& \rotatebox{90}{Aero} & \rotatebox{90}{Bike}&\rotatebox{90}{Bird}&\rotatebox{90}{Boat}&\rotatebox{90}{Bottle}&\rotatebox{90}{Bus}&\rotatebox{90}{Car}&\rotatebox{90}{Cat}&\rotatebox{90}{Chair}&\rotatebox{90}{Cow}&\rotatebox{90}{Table}&\rotatebox{90}{Dog}&\rotatebox{90}{Horse}&\rotatebox{90}{Motor}&\rotatebox{90}{Person}&\rotatebox{90}{Plant}&\rotatebox{90}{Sheep}&\rotatebox{90}{Sofa}&\rotatebox{90}{Train}&\rotatebox{90}{Tv}&mAP\\ 
\midrule
UaDAN~\cite{UaDAN} & 35.0 & 73.7 & 41.0 & 24.4 & 21.3 & 69.8 & 53.5 & 2.3 & 34.2 & 61.2 & 31.0 & \textbf{29.5} & 47.9 & 63.6 & 62.2 & 61.3 & 13.9 & 7.6 & 48.6 & 23.9 & 40.2 \\
FGRR~\cite{FGRR} & 30.8 & 52.1 & 35.1 & 32.4 & 42.2 & 62.8 & 42.6 & 21.4 & 42.8 & 58.6 & 33.5 & 20.8 & 37.2 & 81.4 & 66.2 & 50.3 & 21.5 & 29.3 & \textbf{58.2} & 47.0 & 43.3 \\
UMT~\cite{UMT} & 39.6 & 59.1 & 32.4 & 35.0 & 45.1 & 61.9 & 48.4 & 7.5 & 46.0 & \textbf{67.6} & 21.4 & \textbf{29.5} & 48.2 & 75.9 & 70.5 & \textbf{56.7} & 25.9 & 28.9 & 39.4 & 43.6 & 44.1 \\
SIGMA~\cite{SIGMA} & 40.1 & 55.4 & 37.4 & 31.1 & 54.9 & 54.3 & 46.6 & 23.0 & 44.7 & 65.6 & 23.0 & 22.0 & 42.8 & 55.6 & 67.2 & 55.2 & 32.9 & \textbf{40.8} & 45.0 & 58.6 & 44.5 \\
TIA~\cite{TIA} & 42.2 & 66.0 & 36.9 & 37.3 & 43.7 & \textbf{71.8} & 49.7 & 18.2 & 44.9 & 58.9 & 18.2 & 29.1 & 40.7 & \textbf{87.8} & 67.4 & 49.7 & 27.4 & 27.8 & 57.1 & 50.6 & 46.3 \\
SIGMA++~\cite{sigma++} & 36.3 & 54.6 & 40.1 & 31.6 & 58.0 & 60.4 & 46.2 & \textbf{33.6} & 44.4 & 66.2 & 25.7 & 25.3 & 44.4 & 58.8 & 64.8 & 55.4 & 36.2 & 38.6 & 54.1 & \textbf{59.3} & 46.7 \\
CMT~\cite{CMT} & 39.8 & 56.3 & 38.7 & 39.7 & \textbf{60.4} & 35.0 & \textbf{56.0} & 7.1 & \textbf{60.1} & 60.4 & \textbf{35.8} & 28.1 & \textbf{67.8} & 84.5 & \textbf{80.1} & 55.5 & 20.3 & 32.8 & 42.3 & 38.2 & 47.0 \\
\midrule
DA-Ada(Ours)* & \textbf{42.3} & \textbf{75.1} & \textbf{48.9} & \textbf{45.9} & 49.0 & \textbf{71.8} & 55.6 & 15.4 & 50.7 & 56.6 & 19.9 & 20.6 & 61.3 & 80.7 & 73.0 & 29.2 & \textbf{37.5} & 21.5 & 52.5 & 52.9 & \textbf{48.0} \\
\bottomrule
\end{tabular}  
}
\vspace{-15pt}
\end{table*}

\subsection{Comparison to SOTA methods}
We present representative state-of-the-art DAOD approaches for comparison, including feature alignment and semi-supervised learning methods.

\noindent\textbf{Cross-Weather Adaptation Scenario}
Table~\ref{tab1} (C$\rightarrow$F) illustrates that the proposed DA-Ada surpasses SOTA DA-Pro~\cite{DA-Pro} by a remarkable margin of $2.6\%$, achieving the highest mAP over eight classes of $58.5\%$. 
Compared with existing methods, DA-Ada significantly improves seven categories (\ie ~person, rider, car, truck, bus, train, and bicycle) ranging from $0.4\%$ to $5.3\%$.
The superior performance shows the remarkable effectiveness of the DA-Ada in the cross-domain generalization ability.

\begin{table}[t]
\centering
\Large
\begin{minipage}[t]{0.45\textwidth}
\centering
\caption{\small Comparison ($\%$) of domain-aware adapter with standard adapter.
}
\label{tab-vs}
\resizebox{1.02\columnwidth}{8mm}{
\begin{tabular}{ccccc}
\toprule
Adapter & Source-only& Domain-agnostic& Domain-aware\\ 
\midrule
Cross-Weather  & 50.4 & 53.8 & \textbf{57.1}  \\ 
Cross-FoV      & 57.9 & 63.2 & \textbf{65.8}  \\
Sim-to-Real    & 58.4 & 62.4 & \textbf{65.3}  \\
Cross-Style    & 38.3 & 44.0 & \textbf{46.4}  \\
\bottomrule
\end{tabular} }   
\end{minipage}\hspace{6pt}%
\begin{minipage}[t]{0.52\textwidth}
\centering
\caption{\small Ablation studies ($\%$) of domain-aware adapter on Cross-Weather adaptation.}
\label{tab3}
\setlength\tabcolsep{16pt}
\resizebox{0.95\columnwidth}{8mm}{      
\begin{tabular}{cccccc}
\toprule
DIA & $\mathscr{L}_{dia}$ & DSA & $\mathscr{L}_{dec}$ & mAP & Gains\\ 
\midrule
 &   &    & &52.6 & -\\
 $\checkmark$   &   &  & &53.8 & +1.2\\
 $\checkmark$   &  $\checkmark$ &  & &54.8 & +2.2\\
  $\checkmark$   &  $\checkmark$ & $\checkmark$ & &56.2 & +3.6\\
 $\checkmark$   &  $\checkmark$ & $\checkmark$ & $\checkmark$ &\textbf{57.1} & \textbf{+4.5}\\
\bottomrule
\end{tabular}  }  
\end{minipage}
\vspace{-10pt}
\end{table}


\noindent\textbf{Cross-FOV Adaptation Scenario}
Table~\ref{tab1} (K$\rightarrow$C) indicates a noticeable $5.3\%$ improvement on the SOTA DA-Pro~\cite{DA-Pro} by the DA-Ada.
As K$\rightarrow$C adaptation faces more complicated shape confusion than C$\rightarrow$F, it requests higher discriminability of the model. 
Therefore, the considerable enhancement validates that the DA-Ada can efficiently learn robust visual encoder.

\noindent\textbf{Sim-to-Real Adaptation Scenario}
We report the experimental results on SIM10k $\rightarrow$ Cityscapes benchmark in Table~\ref{tab1} (S$\rightarrow$C).
The proposed DA-Ada achieves the best results of $67.3\%$ mAP, outperforming the previous best entry HT~\cite{HT} $65.5\%$ with $1.8\%$.
The performance of DA-Ada is superior in the difficult adaptation task, which further demonstrates that our strategy is robust not only in appearance but also in more complex semantics adaptation tasks.

\noindent\textbf{Cross-Style Adaptation Scenario}
Additionally, we assess DA-Ada on the more challenging Cross-Style adaptation, where the semantic hierarchy has a broader domain gap. 
DA-Ada peaks with $48.0\%$, outperforming all the SOTA methods presented in Table~\ref{tab2}, demonstrating that injecting cross-domain information into the visual encoder could benefit the adaptation.
Especially, DA-Ada exceeds all the compared methods on six categories (aeroplane, bike, bird, boat, bus, and sheep), which verifies the method is effective under challenging domain shifts and in multi-class problem scenarios.
\vspace{-6pt}

\subsection{Ablation Studies}


\noindent\textbf{Standard Adapter \textit{vs.} Domain-aware Adapter}
We first compare the performance of the domain-aware adapter with existing adapters, including source-only adapter and domain-agnostic adapter.
As shown in Table~\ref{tab-vs}, while the domain-agnostic adapter surpasses the source-only version by $3.4\%\sim 5.7\%$ on four benchmarks, applying the domain-aware adapter further improves $2.4\%\sim 3.3\%$ mAP.
We further explored the reasons for this advantage, shown in Fig~\ref{fig1}(d).
Compared with oracle, the domain-agnostic adapter reaches similar performance on the source domain, but suffers severe performance drop of $3.7\%$ on the target domain, indicating that it is biased towards the source domain.
While improving the source domain with $0.4\%$, our method reaches the oracle on the target domain.
The superioir performance indicates the domain-aware adapter not only aligns domain-invariant knowledge more accurately, but also utilizes domain-specific knowledge to improve the detector's discriminative ability on the target domain.

\begin{table}[t]
    \centering
\large
\begin{minipage}[t]{0.45\textwidth}
\caption{\small Ablation ($\%$) on insertion site of domain-aware adapter on Cross-Weather adaptation. 
}
\label{tab5}
\setlength\tabcolsep{13pt}
\resizebox{0.97\columnwidth}{12mm}{      
\begin{tabular}{ccccc}
\toprule
Block 1 & Block 2 & Block 3 & Block 4 & mAP\\ 
\midrule
 &   &   & $\checkmark$ &  53.6\\
 &   & $\checkmark$ &   &  54.0\\
 & $\checkmark$ &   &   &  54.6\\
 $\checkmark$&  &   &   &  55.1\\
 $\checkmark$& $\checkmark$ &   &   & 56.9\\
 $\checkmark$& $\checkmark$ & $\checkmark$ &   &  57.7\\
 $\checkmark$& $\checkmark$ & $\checkmark$ & $\checkmark$ &  \textbf{58.5}\\
\bottomrule
\end{tabular}  }  
\end{minipage}\hspace{5pt}%
\begin{minipage}[t]{0.52\textwidth}
\caption{\small Ablation ($\%$) on input and injection operation of domain-aware adapter on Cross-Weather adaptation. 
}
\label{tab4}
\setlength\tabcolsep{4.5pt}
\resizebox{0.98\columnwidth}{12mm}{      
\begin{tabular}{cllc}
\toprule
 Input of DIA & Input of DSA & Injection Operation & mAP \\ 
  \midrule
  &  & $\mathscr{F}_i(\mathbf{h}_{i-1})$ & 52.6\\
  $\mathbf{h}_{i-1}$ &        & $\mathbf{h}_i^I=\mathscr{F}_i(\mathbf{h}_{i-1})$ + $\mathscr{A}_i(\mathbf{h}_{i-1})$ & 54.8 \\
  $\mathbf{h}_{i-1}$ & $\mathscr{F}_i(\mathbf{h}_{i-1})$ & $\mathbf{h}_i^I$ + $\mathbf{h}_i^S$& 55.2 \\
  $\mathbf{h}_{i-1}$ & $\mathbf{h}_{i-1}$ - $\mathscr{F}_i(\mathbf{h}_{i-1})$ & $\mathbf{h}_i^I$ + $\mathbf{h}_i^S$& 56.2 \\
  $\mathbf{h}_{i-1}$ & $\mathbf{h}_{i-1}$ - $\mathbf{h}_i^I$ & $\mathbf{h}_i^I$ + $\mathbf{h}_i^S$& 56.7 \\
  $\mathbf{h}_{i-1}$ & $\mathbf{h}_{i-1}$ - $\mathbf{h}_i^I$ & Cross-Attention($\mathbf{h}_i^I$, $\mathbf{h}_i^S$, $\mathbf{h}_i^S$)& 57.0 \\
    \midrule
  $\mathbf{h}_{i-1}$ & $\mathbf{h}_{i-1}$ - $\mathbf{h}_i^I$ & $\mathbf{h}_i^I$ + $\mathbf{h}_i^I \cdot \mathbf{h}_i^S$& \textbf{57.1} \\
\bottomrule
\end{tabular}  }  
\end{minipage}
\vspace{-10pt}
\end{table}

\begin{table}[t!]
    \centering
\large
\begin{minipage}[t]{0.46\textwidth}
\centering
\caption{\small Ablation ($\%$) on Bottleneck dimension of domain-aware adapter. * denotes input dimension.
}
\label{tab6}
\setlength\tabcolsep{8.5pt}
\resizebox{1.\columnwidth}{8.5mm}{      
\begin{tabular}{ccccc}
\toprule
\multicolumn{4}{c}{Bottleneck Dimension} &  \multirow{2}*{mAP}\\
\cmidrule{1-4}
DA-Ada 1 & DA-Ada 2 & DA-Ada 3 & DA-Ada 4 &\\ 
\midrule
 16 & 64  & 128 & 256  & 55.9 \\
 32 & 128 & 256 & 512  & \textbf{57.1} \\
 48 & 192 & 384 & 768  & 56.8 \\
 64* & 256* & 512* & 1024* & 56.6 \\
\bottomrule
\end{tabular}  }  
\end{minipage}\hspace{5pt}%
\begin{minipage}[t]{.5\textwidth}
\caption{\small Comparison ($\%$) of VTA with plain textual tuning methods.
}
\label{tab-vta}
\setlength\tabcolsep{7pt}
\resizebox{0.95\columnwidth}{8.5mm}{      
\begin{tabular}{ccccc}
\toprule
Methods & C$\rightarrow$F & Gains & K$\rightarrow$C & Gains\\ 
\midrule
 Hand-crafted Prompt~\cite{Regionclip} & 52.6  & -     & 59.5 & -    \\
 COOP~\cite{CoOp}                       & 53.5  & +0.9  & 60.7 & +1.2 \\
 DA-Pro~\cite{DA-Pro}                   & 55.1  & +2.5  & 61.4 & +1.9 \\
 VTA(Ours)                              & \textbf{55.8}  & \textbf{+3.2}  & \textbf{62.9} & \textbf{+3.4} \\
\bottomrule
\end{tabular}  }  
\end{minipage}
  \vspace{-15pt} 
\end{table}

\noindent\textbf{Ablation for Domain-aware Adapter}
We conduct comprehensive ablation studies on each component of the proposed method in Table~\ref{tab3}.
Only introducing DIA to the backbone attains an mAP of $53.8\%$, and optimizing each DIA with an independent discriminator $\mathscr{L}_{dia}$ increases $1.0\%$.
This indicates that learning domain-invariant adapters transfer task-related source knowledge to the target domain.
Moreover, the DSA boosts the DIA by $1.4\%$ and $2.3\%$ with the help of $\mathscr{L}_{dec}$, showing that learning domain-specific knowledge improves the discrimination of the target detection head.

\noindent\textbf{Insertion Site}
We explicitly study the insertion site of DA-Ada, as shown in Table~\ref{tab5}.
When single adapter is applied, inserting the DA-Ada in the shallow block achieves better performance, \eg DA-Ada with block 1 obtain $55.1\%$, surpassing all other insertion sites with block 2/3/4.
And increasing the number of DA-Ada from $1$ to $4$ leads to steady improvements of $1.8\%, 0.8\%, 0.8\%$ respectively.

\noindent\textbf{Input and Injection Operation}
We analyze different input features and injection operations of DIA/DSA in Table~\ref{tab4}.
Directly inserting DIA into the visual encoder and directly adding to the output of each block attains $2.2\%$ improvement, showing the effectiveness of learning domain-invariant knowledge.
However, there is limited performance gain in sending the output $\mathscr{F}_i(\mathbf{h}_{i-1})$ to DSA.
It indicates that domain-specific knowledge is ignored during feature extraction of the visual encoder.
To this end, inputting $\mathbf{h}_{i-1} - \mathscr{F}_i(\mathbf{h}_{i-1})$ to DSA receives $56.2\%$, exhibiting that the DSA can regain the domain-specific knowledge from the difference.
As $\mathbf{h}_i^I = \mathscr{A}_i^I(\mathbf{h}_{i-1})+\mathscr{F}_i(\mathbf{h}_{i-1})$ is updated to be domain-invariant, $\mathbf{h}_{i-1}$ - ($\mathbf{h}_i^I$) removes domain-invariant parts and appears to be domain-specific.
Therefore, we forward it to DSA and gain an improvement of $0.5\%$, demonstrating the efficacy of learning domain-specific knowledge.
Additionally, we substitute cross-attention and pixel-level attention for the direct addition, and gains highest mAP of $57.0\%$ and $57.1\%$.
It reveals that domain-specific knowledge describes intra-domain properties and is more suitable for refining the extracted features.
For efficiency, we adopt the simpler pixel-level attention as the fusion function.

\noindent\textbf{Bottleneck Dimension}
We also conduct an ablation study in Table~\ref{tab6} to explore the optimal bottleneck dimension of the DA-Ada. 
As the dimension increases, the performance peaks $57.1\%$ when the bottleneck dimension is $1/2$ of the input and then appears to decline.
We conclude that appropriate dimensional reduction can filter redundant features while extracting task knowledge.

\noindent\textbf{Textual Tuning \textit{vs.} Visual-guided Textual Adapter}
We compare the visual-guided textual adapter against existing methods, as shown in Table~\ref{tab-vta}.
Guided by visual conditions, VTA outperforms SOTA plain textual tuning methods by margins of $0.7\%$ and $1.5\%$ in two scenarios.
Notably, VTA excels in the challenging Cross-FoV adaptation, suggesting that the visual modality effectively supplements the limitations of the textual encoder in describing domain attributes.

\begin{table}[t]
    \centering
\large
\begin{minipage}[t]{.36\textwidth}
\caption{\small Ablation studies ($\%$) of VTA on Cross-Weather adaptation.
}
\label{tab-dita}
\setlength\tabcolsep{8pt}
\resizebox{0.95\columnwidth}{8mm}{      
\begin{tabular}{ccccc}
\toprule
DITA & $\mathscr{L}_{dita}$ & DSTA & mAP & Gains\\ 
\midrule
 &   &   &  57.1 & -\\
 $\checkmark$   &   &  & 57.6 & +0.5\\
 $\checkmark$   &  $\checkmark$ &  &  57.9 & +0.8\\
 $\checkmark$   &  $\checkmark$ & $\checkmark$ & \textbf{58.5} & \textbf{+1.4}\\
\bottomrule
\end{tabular}  }  
\end{minipage}\hspace{6pt}%
\begin{minipage}[t]{0.62\textwidth}
\centering
\captionof{table}{\small Comparison ($\%$) of computational efficiency on Cross-Weather adaptation}
\label{tab-ce}
\setlength\tabcolsep{2pt}
\resizebox{1\columnwidth}{9mm}{      
\begin{tabular}{ccccc}
\toprule
Method & Backbone Param (M) & Learnable Param (M) & mAP & Abs. Gains\\ 
\midrule
DSS~\cite{DSS}        & 29.812 & 29.812  &  40.9   & +4.2  \\ 
CSDA~\cite{CSDA}      & 33.645 & 33.645  &  45.3   & +6.9  \\
AT~\cite{AT}          & \textbf{39.225} & 18.723  &  50.9   & +7.9  \\
DA-Pro~\cite{DA-Pro}  & 34.834 & \textbf{0.008}   &  55.9   & +3.3  \\
DA-Ada(Ours)          & 36.620 & 1.794   &  \textbf{58.5}   & \textbf{+8.0}  \\
\bottomrule
\end{tabular} }   
\end{minipage}
\vspace{-6pt}
\end{table}

\noindent\textbf{Ablation for Visual-guided Textual Adapter}
As shown in Table~\ref{tab-dita}, learning DITA attains an mAP of $57.6\%$, and by introducing an additional adversarial loss, it achieves $57.9\%$.
Moreover, with DSTA generating prompts for each domain, it exhibits a full adaptation performance of $58.5\%$.
This shows that embedding image conditions into textual encoder can promote cross-domain detection.

\noindent\textbf{Computational Efficiency}
As shown in Table~\ref{tab-ce}, employing VLM yields a similar parameter scale while achieving a peak mAP of $58.5\%$. 
This indicates that the superior performance of DA-Ada does not arise from an increase in parameters. Furthermore, DA-Ada achieves the highest absolute gain of $+8.0\%$ with the training of only 1.794M parameters, demonstrating remarkable efficiency.

\vspace{-5pt} 
\subsection{Detection Visualization}
\begin{figure*}[t]
  \centering
  \includegraphics[width=0.96\textwidth]{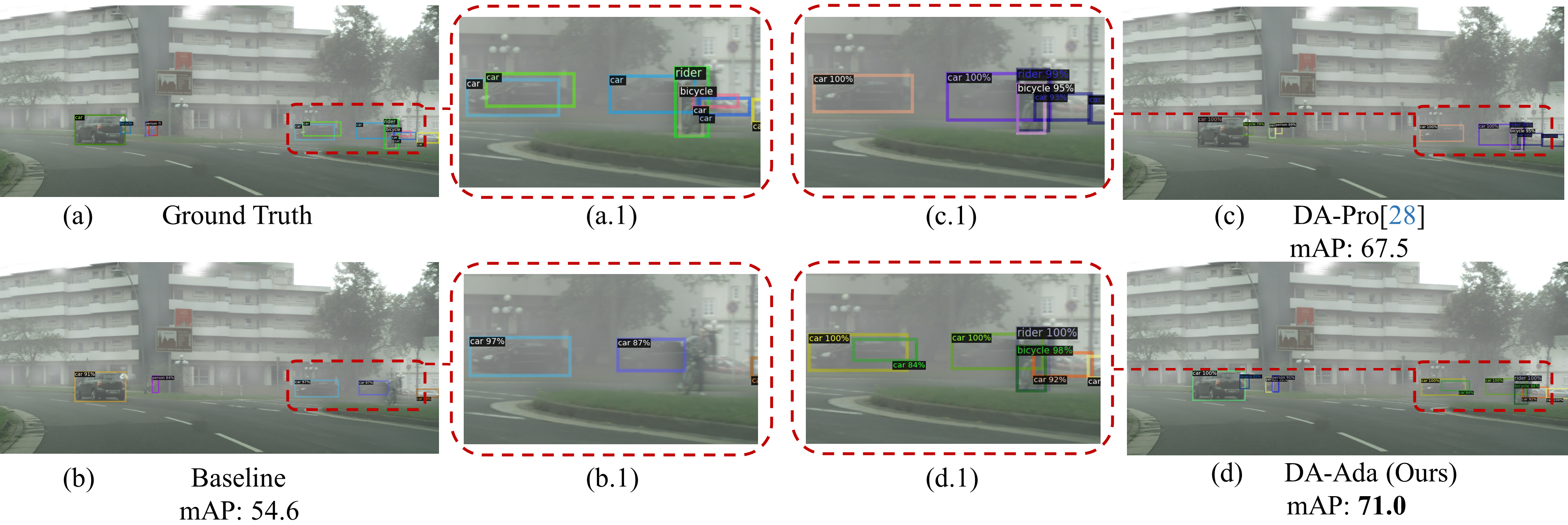}
  \vspace{-2pt}
  \caption{Detection comparison on the Cross-Weather adaptation scenario. 
  We visualize the ground truth (a), the detection boxes of SOTA DA-Pro~\cite{DA-Pro}(c) and our methods (b)(d).
  mAP: mean Average Precision on the example image}
  \label{fig3}
  \vspace{-10pt}
\end{figure*}

In Fig.~\ref{fig3}, we present the comparison of the ground truth boxes (a) and the detection boxes of SOTA DA-Pro~\cite{DA-Pro}(c) and our methods (b)(d) on the target domain.
(a.1)(b.1)(c.1)(d.1) are zoomed from the same region of images (a)(b)(c)(d) for a better view.
Fig.~\ref{fig3}(a.1) presents eight objects in the cropped region: 6 overlapped cars and a rider with a bicycle.
The baseline model only detects two clear cars in Fig.~\ref{fig3}(b.1).
Failing to describe domain information, like weather conditions, it misses other objects hidden in the fog.
In Fig.~\ref{fig3}(c), the DA-Pro distinguishes the rider and the bicycle and improves $9.3\%$ mAP with the domain-adaptive prompt.
However, it ignores one car on the left of Fig.~\ref{fig3}(c.1), suffering limited generalization ability due to insufficient domain representation learning in the visual encoder.
Our proposed DA-Ada correctly detects the missing car (labelled in green) in the cropped region Fig.~\ref{fig3}(d.1).
By injecting cross-domain information into the visual encoder, the DA-Ada enables the model to detect more confidently on two bicycles ($83\%, 98\%$) and one person ($91\%$), compared with DA-Pro's ($79\%, 95\%$) and ($89\%$).
These comparison results reveal the effectiveness of DA-Ada.
\vspace{-5pt}

\subsection{Feature Visualization}
\begin{figure*}[t]
  \centering
  \includegraphics[width=1.0\textwidth]{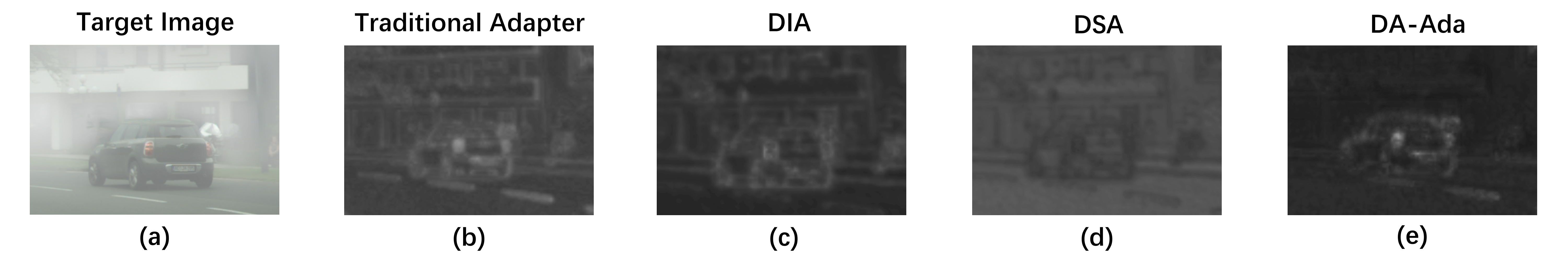}
  \vspace{-2pt}
  \caption{Feature comparison on the Cross-Weather adaptation scenario. We visualize (a) the target image and the output
feature of (b) traditional adapter, (b) domain-invariant adapter (DIA), (c) domain-specific adapter (DSA) and (d) domain-aware adapter (DA-Ada).}
  \label{fig4}
  \vspace{-10pt}
\end{figure*}

In Fig.~\ref{fig4}, we visualize the output features of the traditional adapter, the domain-invariant adapter (DIA), domain-specific adapter (DSA) and the domain-aware adapter (DA-Ada). 
We sample image (a) a car and a person in the fog from Foggy Cityscapes. 
The traditional adapter (b) roughly extracts the outline of the car. However, affected by target domain attributes, such as fog, background areas are also highlighted in (b), and the person is not salient. 
DIA (C) mainly focuses on the object area and extracts domain-shared task information. 
DSA (d) mainly focuses on factors related to domain attributes besides the objects, such as foggy areas. 
By combining DIA with DSA, DA-Ada (e) extracts the car and person while reducing the interference of fog in the background. 
Compared with (b), objects are more salient in (e), indicating the effectiveness of DA-Ada.

\vspace{-5pt} 
\section{Conclusion}
\vspace{-5pt}
In this paper, we propose a novel Domain-Aware Adapter (DA-Ada) for DAOD.
As a small learnable attachment, it transfers highly generalized knowledge the visual-language model provides to cross-domain information for DAOD.
Precisely, it consists of a Domain-Invariant Adapter (DIA) for learning domain-invariant knowledge and a Domain-Specific Adapter (DSA) for recovering the domain-specific knowledge from information discarded by the visual encoder.
Extensive experiments over multiple DAOD tasks validate the effectiveness of DA-Ada in inferring a discriminative detector.


{\small
\bibliographystyle{plainnat}
\bibliography{egbib.bib}

\begin{thebibliography}{82}
\providecommand{\natexlab}[1]{#1}
\providecommand{\url}[1]{\texttt{#1}}
\expandafter\ifx\csname urlstyle\endcsname\relax
  \providecommand{\doi}[1]{doi: #1}\else
  \providecommand{\doi}{doi: \begingroup \urlstyle{rm}\Url}\fi

\bibitem[Bousmalis et~al.(2016)Bousmalis, Trigeorgis, Silberman, Krishnan, and Erhan]{DSN}
Konstantinos Bousmalis, George Trigeorgis, Nathan Silberman, Dilip Krishnan, and Dumitru Erhan.
\newblock Domain separation networks.
\newblock In \emph{NeurIPS}, pages 343--351, 2016.

\bibitem[Chang et~al.(2019)Chang, You, Seo, Kwak, and Han]{DSB}
Woong-Gi Chang, Tackgeun You, Seonguk Seo, Suha Kwak, and Bohyung Han.
\newblock Domain-specific batch normalization for unsupervised domain adaptation.
\newblock In \emph{CVPR}, pages 7354--7362, 2019.

\bibitem[Chen et~al.(2021)Chen, Li, Zheng, Huang, Ding, and Yu]{DBGL}
Chaoqi Chen, Jiongcheng Li, Zebiao Zheng, Yue Huang, Xinghao Ding, and Yizhou Yu.
\newblock Dual bipartite graph learning: A general approach for domain adaptive object detection.
\newblock In \emph{ICCV}, pages 2703--2712, 2021.

\bibitem[Chen et~al.(2023)Chen, Li, Zhou, Han, Huang, Ding, and Yu]{FGRR}
Chaoqi Chen, Jiongcheng Li, Hong-Yu Zhou, Xiaoguang Han, Yue Huang, Xinghao Ding, and Yizhou Yu.
\newblock Relation matters: foreground-aware graph-based relational reasoning for domain adaptive object detection.
\newblock \emph{TPAMI}, 45\penalty0 (03):\penalty0 3677--3694, 2023.

\bibitem[Chen et~al.(2022{\natexlab{a}})Chen, Tao, Zhang, Wang, Ye, Wang, Hu, and Savvides]{Conv-adapter}
Hao Chen, Ran Tao, Han Zhang, Yidong Wang, Wei Ye, Jindong Wang, Guosheng Hu, and Marios Savvides.
\newblock Conv-adapter: Exploring parameter efficient transfer learning for convnets.
\newblock \emph{arXiv preprint arXiv:2208.07463}, 2022{\natexlab{a}}.

\bibitem[Chen et~al.(2022{\natexlab{b}})Chen, Chen, Yang, Song, Wang, Zhang, Yan, Qi, Zhuang, Xie, et~al.]{PT}
Meilin Chen, Weijie Chen, Shicai Yang, Jie Song, Xinchao Wang, Lei Zhang, Yunfeng Yan, Donglian Qi, Yueting Zhuang, Di~Xie, et~al.
\newblock Learning domain adaptive object detection with probabilistic teacher.
\newblock In \emph{ICML}, pages 3040--3055, 2022{\natexlab{b}}.

\bibitem[Chen et~al.(2018)Chen, Li, Sakaridis, Dai, and Van~Gool]{DA-Faster}
Yuhua Chen, Wen Li, Christos Sakaridis, Dengxin Dai, and Luc Van~Gool.
\newblock Domain adaptive faster r-cnn for object detection in the wild.
\newblock In \emph{CVPR}, pages 3339--3348, 2018.

\bibitem[Chen et~al.(2022{\natexlab{c}})Chen, Duan, Wang, He, Lu, Dai, and Qiao]{V-adapter}
Zhe Chen, Yuchen Duan, Wenhai Wang, Junjun He, Tong Lu, Jifeng Dai, and Yu~Qiao.
\newblock Adapterfusion: Non-destructive task composition for transfer learning.
\newblock \emph{arXiv preprint arXiv:2205.08534}, 2022{\natexlab{c}}.

\bibitem[Cordts et~al.(2016)Cordts, Omran, Ramos, Rehfeld, Enzweiler, Benenson, Franke, Roth, and Schiele]{Cityscapes}
Marius Cordts, Mohamed Omran, Sebastian Ramos, Timo Rehfeld, Markus Enzweiler, Rodrigo Benenson, Uwe Franke, Stefan Roth, and Bernt Schiele.
\newblock The cityscapes dataset for semantic urban scene understanding.
\newblock In \emph{CVPR}, 2016.

\bibitem[Deng et~al.(2021)Deng, Li, Chen, and Duan]{UMT}
Jinhong Deng, Wen Li, Yuhua Chen, and Lixin Duan.
\newblock Unbiased mean teacher for cross-domain object detection.
\newblock In \emph{CVPR}, pages 4091--4101, 2021.

\bibitem[Deng et~al.(2023)Deng, Xu, Li, and Duan]{HT}
Jinhong Deng, Dongli Xu, Wen Li, and Lixin Duan.
\newblock Harmonious teacher for cross-domain object detection.
\newblock In \emph{CVPR}, pages 23829--23838, 2023.

\bibitem[Dosovitskiy et~al.(2021)Dosovitskiy, Beyer, Kolesnikov, Weissenborn, Zhai, Unterthiner, Dehghani, Minderer, Heigold, Gelly, Uszkoreit, and Houlsby]{ViT}
Alexey Dosovitskiy, Lucas Beyer, Alexander Kolesnikov, Dirk Weissenborn, Xiaohua Zhai, Thomas Unterthiner, Mostafa Dehghani, Matthias Minderer, Georg Heigold, Sylvain Gelly, Jakob Uszkoreit, and Neil Houlsby.
\newblock An image is worth 16x16 words: Transformers for image recognition at scale.
\newblock In \emph{ICLR}, 2021.

\bibitem[Everingham et~al.(2015)Everingham, Eslami, Van~Gool, Williams, Winn, and Zisserman]{pascal}
Mark Everingham, SM~Ali Eslami, Luc Van~Gool, Christopher~KI Williams, John Winn, and Andrew Zisserman.
\newblock The pascal visual object classes challenge: A retrospective.
\newblock \emph{IJCV}, 111:\penalty0 98--136, 2015.

\bibitem[Fahes et~al.(2023)Fahes, Vu, Bursuc, P{\'e}rez, and De~Charette]{poda}
Mohammad Fahes, Tuan-Hung Vu, Andrei Bursuc, Patrick P{\'e}rez, and Raoul De~Charette.
\newblock Poda: Prompt-driven zero-shot domain adaptation.
\newblock In \emph{ICCV}, pages 18623--18633, 2023.

\bibitem[Ganin et~al.(2016)Ganin, Ustinova, Ajakan, Germain, Larochelle, Laviolette, Marchand, and Lempitsky]{DANN}
Yaroslav Ganin, Evgeniya Ustinova, Hana Ajakan, Pascal Germain, Hugo Larochelle, Fran{\c{c}}ois Laviolette, Mario Marchand, and Victor Lempitsky.
\newblock Domain-adversarial training of neural networks.
\newblock \emph{JMLR}, 17\penalty0 (1):\penalty0 2096--2030, 2016.

\bibitem[Gao et~al.(2023)Gao, Liu, Dun, and Qian]{CSDA}
Changlong Gao, Chengxu Liu, Yujie Dun, and Xueming Qian.
\newblock Csda: Learning category-scale joint feature for domain adaptive object detection.
\newblock In \emph{ICCV}, pages 11421--11430, 2023.

\bibitem[Gao et~al.(2024)Gao, Geng, Zhang, Ma, Fang, Zhang, Li, and Qiao]{CLIP-Adapter}
Peng Gao, Shijie Geng, Renrui Zhang, Teli Ma, Rongyao Fang, Yongfeng Zhang, Hongsheng Li, and Yu~Qiao.
\newblock Clip-adapter: Better vision-language models with feature adapters.
\newblock \emph{IJCV}, 132\penalty0 (2):\penalty0 581--595, 2024.

\bibitem[Geiger et~al.(2012)Geiger, Lenz, and Urtasun]{KITTI}
Andreas Geiger, Philip Lenz, and Raquel Urtasun.
\newblock Are we ready for autonomous driving? the kitti vision benchmark suite.
\newblock In \emph{CVPR}, 2012.

\bibitem[Gu et~al.(2022)Gu, Lin, Kuo, and Cui]{ViLD}
Xiuye Gu, Tsung-Yi Lin, Weicheng Kuo, and Yin Cui.
\newblock Open-vocabulary object detection via vision and language knowledge distillation.
\newblock In \emph{ICLR}, 2022.

\bibitem[Guan et~al.(2021)Guan, Huang, Xiao, Lu, and Cao]{UaDAN}
Dayan Guan, Jiaxing Huang, Aoran Xiao, Shijian Lu, and Yanpeng Cao.
\newblock Uncertainty-aware unsupervised domain adaptation in object detection.
\newblock \emph{TMM}, 24:\penalty0 2502--2514, 2021.

\bibitem[He et~al.(2021)He, Zhou, Ma, Berg-Kirkpatrick, and Neubig]{unified}
Junxian He, Chunting Zhou, Xuezhe Ma, Taylor Berg-Kirkpatrick, and Graham Neubig.
\newblock Towards a unified view of parameter-efficient transfer learning.
\newblock In \emph{ICLR}, 2021.

\bibitem[He et~al.(2016)He, Zhang, Ren, and Sun]{Resnet}
Kaiming He, Xiangyu Zhang, Shaoqing Ren, and Jian Sun.
\newblock Deep residual learning for image recognition.
\newblock In \emph{CVPR}, pages 770--778, 2016.

\bibitem[He et~al.(2022)He, Wang, Wu, Wang, Li, Li, Gan, Wu, and Qiao]{TDD}
Mengzhe He, Yali Wang, Jiaxi Wu, Yiru Wang, Hanqing Li, Bo~Li, Weihao Gan, Wei Wu, and Yu~Qiao.
\newblock Cross domain object detection by target-perceived dual branch distillation.
\newblock In \emph{CVPR}, pages 9570--9580, 2022.

\bibitem[Houlsby et~al.(2019)Houlsby, Giurgiu, Jastrzebski, Morrone, De~Laroussilhe, Gesmundo, Attariyan, and Gelly]{Adapter}
Neil Houlsby, Andrei Giurgiu, Stanislaw Jastrzebski, Bruna Morrone, Quentin De~Laroussilhe, Andrea Gesmundo, Mona Attariyan, and Sylvain Gelly.
\newblock Parameter-efficient transfer learning for nlp.
\newblock In \emph{ICML}, pages 2790--2799, 2019.

\bibitem[Huang et~al.(2022)Huang, Guan, Xiao, Lu, and Shao]{Category-contrast}
Jiaxing Huang, Dayan Guan, Aoran Xiao, Shijian Lu, and Ling Shao.
\newblock Category contrast for unsupervised domain adaptation in visual tasks.
\newblock In \emph{CVPR}, pages 1203--1214, 2022.

\bibitem[Inoue et~al.(2018)Inoue, Furuta, Yamasaki, and Aizawa]{clipart}
Naoto Inoue, Ryosuke Furuta, Toshihiko Yamasaki, and Kiyoharu Aizawa.
\newblock Cross-domain weakly-supervised object detection through progressive domain adaptation.
\newblock In \emph{CVPR}, pages 5001--5009, 2018.

\bibitem[Johnson-Roberson et~al.(2017)Johnson-Roberson, Barto, Mehta, Sridhar, Rosaen, and Vasudevan]{SIM10K}
Matthew Johnson-Roberson, Charles Barto, Rounak Mehta, Sharath~Nittur Sridhar, Karl Rosaen, and Ram Vasudevan.
\newblock Driving in the matrix: Can virtual worlds replace human-generated annotations for real world tasks?
\newblock In \emph{ICRA 2017}, pages 746--753. IEEE, 2017.

\bibitem[Kirillov et~al.(2023)Kirillov, Mintun, Ravi, Mao, Rolland, Gustafson, Xiao, Whitehead, Berg, Lo, et~al.]{SAM}
Alexander Kirillov, Eric Mintun, Nikhila Ravi, Hanzi Mao, Chloe Rolland, Laura Gustafson, Tete Xiao, Spencer Whitehead, Alexander~C Berg, Wan-Yen Lo, et~al.
\newblock Segment anything.
\newblock In \emph{ICCV}, pages 4015--4026, 2023.

\bibitem[Li et~al.(2020)Li, Du, Zhang, Wen, Luo, Wu, and Zhu]{SAP}
Congcong Li, Dawei Du, Libo Zhang, Longyin Wen, Tiejian Luo, Yanjun Wu, and Pengfei Zhu.
\newblock Spatial attention pyramid network for unsupervised domain adaptation.
\newblock In \emph{ECCV}, pages 481--497. Springer, 2020.

\bibitem[Li et~al.(2024)Li, Zhang, Yao, Song, Hao, Zhao, Li, and Chen]{DA-Pro}
Haochen Li, Rui Zhang, Hantao Yao, Xinkai Song, Yifan Hao, Yongwei Zhao, Ling Li, and Yunji Chen.
\newblock Learning domain-aware detection head with prompt tuning.
\newblock \emph{NeurIPS}, 36, 2024.

\bibitem[Li et~al.(2022{\natexlab{a}})Li, Li, Xiong, and Hoi]{BLIP}
Junnan Li, Dongxu Li, Caiming Xiong, and Steven Hoi.
\newblock {BLIP}: Bootstrapping language-image pre-training for unified vision-language understanding and generation.
\newblock In \emph{ICML}, pages 12888--12900, 2022{\natexlab{a}}.

\bibitem[Li et~al.(2023{\natexlab{a}})Li, Li, Savarese, and Hoi]{BLIP2}
Junnan Li, Dongxu Li, Silvio Savarese, and Steven Hoi.
\newblock {BLIP-2}: Bootstrapping language-image pre-training with frozen image encoders and large language models.
\newblock \emph{arXiv preprint arXiv:2301.12597}, 2023{\natexlab{a}}.

\bibitem[Li et~al.(2023{\natexlab{b}})Li, Wigington, Tensmeyer, Morariu, Zhao, Manjunatha, Barmpalios, and Fu]{ATMT}
Kai Li, Curtis Wigington, Chris Tensmeyer, Vlad~I. Morariu, Handong Zhao, Varun Manjunatha, Nikolaos Barmpalios, and Yun Fu.
\newblock Improving cross-domain detection with self-supervised learning.
\newblock In \emph{CVPR}, pages 4745--4754, 2023{\natexlab{b}}.

\bibitem[Li et~al.(2022{\natexlab{b}})Li, Liu, Yao, and Yuan]{SCAN}
Wuyang Li, Xinyu Liu, Xiwen Yao, and Yixuan Yuan.
\newblock Scan: Cross domain object detection with semantic conditioned adaptation.
\newblock In \emph{AAAI}, volume~6, page~7, 2022{\natexlab{b}}.

\bibitem[Li et~al.(2022{\natexlab{c}})Li, Liu, and Yuan]{SIGMA}
Wuyang Li, Xinyu Liu, and Yixuan Yuan.
\newblock Sigma: Semantic-complete graph matching for domain adaptive object detection.
\newblock In \emph{CVPR}, pages 5291--5300, 2022{\natexlab{c}}.

\bibitem[Li et~al.(2023{\natexlab{c}})Li, Liu, and Yuan]{sigma++}
Wuyang Li, Xinyu Liu, and Yixuan Yuan.
\newblock Sigma++: Improved semantic-complete graph matching for domain adaptive object detection.
\newblock \emph{TPAMI}, 45\penalty0 (07):\penalty0 9022--9040, 2023{\natexlab{c}}.

\bibitem[Li et~al.(2022{\natexlab{d}})Li, Dai, Ma, Liu, Chen, Wu, He, Kitani, and Vajda]{AT}
Yu-Jhe Li, Xiaoliang Dai, Chih-Yao Ma, Yen-Cheng Liu, Kan Chen, Bichen Wu, Zijian He, Kris Kitani, and Peter Vajda.
\newblock Cross-domain adaptive teacher for object detection.
\newblock In \emph{CVPR}, pages 7581--7590, 2022{\natexlab{d}}.

\bibitem[Lin et~al.(2021)Lin, Yuan, Zhao, Sun, Wang, and Cai]{DIDN}
Chuang Lin, Zehuan Yuan, Sicheng Zhao, Peize Sun, Changhu Wang, and Jianfei Cai.
\newblock Domain-invariant disentangled network for generalizable object detection.
\newblock In \emph{ICCV}, pages 8771--8780, 2021.

\bibitem[Lin et~al.(2017)Lin, Doll{\'a}r, Girshick, He, Hariharan, and Belongie]{FPN}
Tsung-Yi Lin, Piotr Doll{\'a}r, Ross Girshick, Kaiming He, Bharath Hariharan, and Serge Belongie.
\newblock Feature pyramid networks for object detection.
\newblock In \emph{CVPR}, pages 2117--2125, 2017.

\bibitem[Liu et~al.(2022)Liu, Zhang, Song, Huang, Wang, Barnett, and Cai]{DDF}
Dongnan Liu, Chaoyi Zhang, Yang Song, Heng Huang, Chenyu Wang, Michael Barnett, and Weidong Cai.
\newblock Decompose to adapt: Cross-domain object detection via feature disentanglement.
\newblock \emph{TMM}, 25:\penalty0 1333--1344, 2022.

\bibitem[Liu et~al.(2021)Liu, Han, Wang, and Tian]{FSAC}
Rui Liu, Yahong Han, Yaowei Wang, and Qi~Tian.
\newblock Frequency spectrum augmentation consistency for domain adaptive object detection.
\newblock \emph{arXiv preprint arXiv:2112.08605}, 2021.

\bibitem[Liu et~al.(2023)Liu, Wang, Huang, Wang, and Xu]{CIGAR}
Yabo Liu, Jinghua Wang, Chao Huang, Yaowei Wang, and Yong Xu.
\newblock Cigar: Cross-modality graph reasoning for domain adaptive object detection.
\newblock In \emph{CVPR}, pages 23776--23786, 2023.

\bibitem[Long et~al.(2015)Long, Cao, Wang, and Jordan]{DAN}
Mingsheng Long, Yue Cao, Jianmin Wang, and Michael Jordan.
\newblock Learning transferable features with deep adaptation networks.
\newblock In \emph{ICML}, pages 97--105. PMLR, 2015.

\bibitem[Long et~al.(2016)Long, Zhu, Wang, and Jordan]{RTN}
Mingsheng Long, Han Zhu, Jianmin Wang, and Michael~I Jordan.
\newblock Unsupervised domain adaptation with residual transfer networks.
\newblock \emph{NeurIPS}, 29, 2016.

\bibitem[Long et~al.(2017)Long, Zhu, Wang, and Jordan]{JAN}
Mingsheng Long, Han Zhu, Jianmin Wang, and Michael~I Jordan.
\newblock Deep transfer learning with joint adaptation networks.
\newblock In \emph{ICML}, pages 2208--2217. PMLR, 2017.

\bibitem[Pantazis et~al.(2022)Pantazis, Brostow, Jones, and Mac~Aodha]{SVL-Adapter}
Omiros Pantazis, Gabriel Brostow, Kate Jones, and Oisin Mac~Aodha.
\newblock Svl-adapter: Self-supervised adapter for vision-language pretrained models.
\newblock \emph{arXiv preprint arXiv:2210.03794}, 2022.

\bibitem[Pfeiffer et~al.(2020)Pfeiffer, Kamath, R{\"u}ckl{\'e}, Cho, and Gurevych]{adapterfusion}
Jonas Pfeiffer, Aishwarya Kamath, Andreas R{\"u}ckl{\'e}, Kyunghyun Cho, and Iryna Gurevych.
\newblock Adapterfusion: Non-destructive task composition for transfer learning.
\newblock \emph{arXiv preprint arXiv:2005.00247}, 2020.

\bibitem[Radford et~al.(2021)Radford, Kim, Hallacy, Ramesh, Goh, Agarwal, Sastry, Askell, Mishkin, Clark, et~al.]{CLIP}
Alec Radford, Jong~Wook Kim, Chris Hallacy, Aditya Ramesh, Gabriel Goh, Sandhini Agarwal, Girish Sastry, Amanda Askell, Pamela Mishkin, Jack Clark, et~al.
\newblock Learning transferable visual models from natural language supervision.
\newblock In \emph{ICML}, pages 8748--8763. PMLR, 2021.

\bibitem[Redmon et~al.(2016)Redmon, Divvala, Girshick, and Farhadi]{YOLO}
Joseph Redmon, Santosh Divvala, Ross Girshick, and Ali Farhadi.
\newblock You only look once: Unified, real-time object detection.
\newblock In \emph{CVPR}, pages 779--788, 2016.

\bibitem[Ren et~al.(2015)Ren, He, Girshick, and Sun]{FasterRCNN}
Shaoqing Ren, Kaiming He, Ross Girshick, and Jian Sun.
\newblock Faster r-cnn: Towards real-time object detection with region proposal networks.
\newblock \emph{NeurIPS}, 28, 2015.

\bibitem[Saito et~al.(2019)Saito, Ushiku, Harada, and Saenko]{Strong-weak}
Kuniaki Saito, Yoshitaka Ushiku, Tatsuya Harada, and Kate Saenko.
\newblock Strong-weak distribution alignment for adaptive object detection.
\newblock In \emph{CVPR}, pages 6956--6965, 2019.

\bibitem[Sakaridis et~al.(2018)Sakaridis, Dai, and Van~Gool]{FoggyCityscapes}
Christos Sakaridis, Dengxin Dai, and Luc Van~Gool.
\newblock Semantic foggy scene understanding with synthetic data.
\newblock \emph{IJCV}, 126\penalty0 (9):\penalty0 973--992, Sep 2018.

\bibitem[Sanyal et~al.(2023)Sanyal, Asokan, Bhambri, Kulkarni, Kundu, and Babu]{DS}
Sunandini Sanyal, Ashish~Ramayee Asokan, Suvaansh Bhambri, Akshay Kulkarni, Jogendra~Nath Kundu, and R~Venkatesh Babu.
\newblock Domain-specificity inducing transformers for source-free domain adaptation.
\newblock In \emph{ICCV}, pages 18928--18937, 2023.

\bibitem[Shao et~al.(2018)Shao, Lan, and Yuen]{FCP}
Rui Shao, Xiangyuan Lan, and Pong~C Yuen.
\newblock Feature constrained by pixel: Hierarchical adversarial deep domain adaptation.
\newblock In \emph{ACMMM}, pages 220--228, 2018.

\bibitem[Singha et~al.(2023)Singha, Pal, Jha, and Banerjee]{AD-Clip}
Mainak Singha, Harsh Pal, Ankit Jha, and Biplab Banerjee.
\newblock Ad-clip: Adapting domains in prompt space using clip.
\newblock In \emph{ICCV}, pages 4355--4364, 2023.

\bibitem[Sung et~al.(2022)Sung, Cho, and Bansal]{VL-adapter}
Yi-Lin Sung, Jaemin Cho, and Mohit Bansal.
\newblock Vl-adapter: Parameter-efficient transfer learning for vision-and-language tasks.
\newblock In \emph{CVPR}, pages 5227--5237, 2022.

\bibitem[Tarvainen and Valpola(2017)]{MT}
Antti Tarvainen and Harri Valpola.
\newblock Mean teachers are better role models: Weight-averaged consistency targets improve semi-supervised deep learning results.
\newblock \emph{NPIS}, 30, 2017.

\bibitem[Vibashan et~al.(2023)Vibashan, Oza, and Patel]{CMT}
VS~Vibashan, Poojan Oza, and Vishal~M Patel.
\newblock Instance relation graph guided source-free domain adaptive object detection.
\newblock In \emph{CVPR}, pages 3520--3530, 2023.

\bibitem[Vidit et~al.(2023)Vidit, Engilberge, and Salzmann]{ClipTheGap}
Vidit Vidit, Martin Engilberge, and Mathieu Salzmann.
\newblock Clip the gap: A single domain generalization approach for object detection.
\newblock In \emph{CVPR}, pages 3219--3229, 2023.

\bibitem[Vs et~al.(2021)Vs, Gupta, Oza, Sindagi, and Patel]{MEGA-CDA}
Vibashan Vs, Vikram Gupta, Poojan Oza, Vishwanath~A Sindagi, and Vishal~M Patel.
\newblock Mega-cda: Memory guided attention for category-aware unsupervised domain adaptive object detection.
\newblock In \emph{CVPR}, pages 4516--4526, 2021.

\bibitem[Wang et~al.(2024)Wang, Jia, Zeng, Zhang, and Li]{TFD}
Haoan Wang, Shilong Jia, Tieyong Zeng, Guixu Zhang, and Zhi Li.
\newblock Triple feature disentanglement for one-stage adaptive object detection.
\newblock In \emph{AAAI}, pages 5401--5409, 2024.

\bibitem[Wang et~al.(2020)Wang, Tang, Duan, Wei, Huang, Cao, Jiang, Zhou, et~al.]{k-adapter}
Ruize Wang, Duyu Tang, Nan Duan, Zhongyu Wei, Xuanjing Huang, Guihong Cao, Daxin Jiang, Ming Zhou, et~al.
\newblock K-adapter: Infusing knowledge into pre-trained models with adapters.
\newblock \emph{arXiv preprint arXiv:2002.01808}, 2020.

\bibitem[Wang et~al.(2021)Wang, Zhang, Zhang, Li, Xia, Zhang, and Liu]{DSS}
Yu~Wang, Rui Zhang, Shuo Zhang, Miao Li, Yangyang Xia, XiShan Zhang, and ShaoLi Liu.
\newblock Domain-specific suppression for adaptive object detection.
\newblock In \emph{CVPR}, pages 9603--9612, 2021.

\bibitem[Wu and Deng(2022)]{CDSD}
Aming Wu and Cheng Deng.
\newblock Single-domain generalized object detection in urban scene via cyclic-disentangled self-distillation.
\newblock In \emph{CVPR}, pages 847--856, 2022.

\bibitem[Wu et~al.(2021{\natexlab{a}})Wu, Han, Zhu, and Yang]{IIPD}
Aming Wu, Yahong Han, Linchao Zhu, and Yi~Yang.
\newblock Instance-invariant domain adaptive object detection via progressive disentanglement.
\newblock \emph{TPAMI}, 44\penalty0 (8):\penalty0 4178--4193, 2021{\natexlab{a}}.

\bibitem[Wu et~al.(2021{\natexlab{b}})Wu, Liu, Han, Zhu, and Yang]{VD}
Aming Wu, Rui Liu, Yahong Han, Linchao Zhu, and Yi~Yang.
\newblock Vector-decomposed disentanglement for domain-invariant object detection.
\newblock In \emph{ICCV}, pages 9342--9351, 2021{\natexlab{b}}.

\bibitem[Wu et~al.(2022)Wu, Chen, He, Wang, Li, Ma, Gan, Wu, Wang, and Huang]{TRKP}
Jiaxi Wu, Jiaxin Chen, Mengzhe He, Yiru Wang, Bo~Li, Bingqi Ma, Weihao Gan, Wei Wu, Yali Wang, and Di~Huang.
\newblock Target-relevant knowledge preservation for multi-source domain adaptive object detection.
\newblock In \emph{CVPR}, pages 5301--5310, 2022.

\bibitem[Wu et~al.(2023)Wu, Ji, Liu, Fu, Xu, Xu, and Jin]{Medical-sam-adapter}
Junde Wu, Wei Ji, Yuanpei Liu, Huazhu Fu, Min Xu, Yanwu Xu, and Yueming Jin.
\newblock Medical sam adapter: Adapting segment anything model for medical image segmentation.
\newblock \emph{arXiv preprint arXiv:2304.12620}, 2023.

\bibitem[Xu et~al.(2020)Xu, Wang, Ni, Tian, and Zhang]{GPA}
Minghao Xu, Hang Wang, Bingbing Ni, Qi~Tian, and Wenjun Zhang.
\newblock Cross-domain detection via graph-induced prototype alignment.
\newblock In \emph{CVPR}, pages 12355--12364, 2020.

\bibitem[Yan et~al.(2017)Yan, Ding, Li, Wang, Xu, and Zuo]{WMMD}
Hongliang Yan, Yukang Ding, Peihua Li, Qilong Wang, Yong Xu, and Wangmeng Zuo.
\newblock Mind the class weight bias: Weighted maximum mean discrepancy for unsupervised domain adaptation.
\newblock In \emph{CVPR}, pages 2272--2281, 2017.

\bibitem[Yao et~al.(2023)Yao, Zhang, and Xu]{KgCoOp}
Hantao Yao, Rui Zhang, and Changsheng Xu.
\newblock Visual-language prompt tuning with knowledge-guided context optimization.
\newblock In \emph{Proceedings of the IEEE/CVF Conference on Computer Vision and Pattern Recognition}, pages 6757--6767, 2023.

\bibitem[Yoo et~al.(2022)Yoo, Chung, and Kwak]{OADA}
Jayeon Yoo, Inseop Chung, and Nojun Kwak.
\newblock Unsupervised domain adaptation for one-stage object detector using offsets to bounding box.
\newblock In \emph{ECCV}, pages 691--708, 2022.

\bibitem[Zhang et~al.(2015)Zhang, Yu, Chang, and Wang]{DTN}
Xu~Zhang, Felix~Xinnan Yu, Shih-Fu Chang, and Shengjin Wang.
\newblock Deep transfer network: Unsupervised domain adaptation.
\newblock \emph{arXiv preprint arXiv:1503.00591}, 2015.

\bibitem[Zhang et~al.(2021)Zhang, Wang, and Mao]{RPN}
Yixin Zhang, Zilei Wang, and Yushi Mao.
\newblock Rpn prototype alignment for domain adaptive object detector.
\newblock In \emph{CVPR}, pages 12425--12434, 2021.

\bibitem[Zhao and Wang(2022)]{TIA}
Liang Zhao and Limin Wang.
\newblock Task-specific inconsistency alignment for domain adaptive object detection.
\newblock In \emph{CVPR}, pages 14217--14226, 2022.

\bibitem[Zhong et~al.(2022)Zhong, Yang, Zhang, Li, Codella, Li, Zhou, Dai, Yuan, Li, et~al.]{Regionclip}
Yiwu Zhong, Jianwei Yang, Pengchuan Zhang, Chunyuan Li, Noel Codella, Liunian~Harold Li, Luowei Zhou, Xiyang Dai, Lu~Yuan, Yin Li, et~al.
\newblock Regionclip: Region-based language-image pretraining.
\newblock In \emph{CVPR}, pages 16793--16803, 2022.

\bibitem[Zhou et~al.(2022{\natexlab{a}})Zhou, Yang, Loy, and Liu]{CoCoOp}
Kaiyang Zhou, Jingkang Yang, Chen~Change Loy, and Ziwei Liu.
\newblock Conditional prompt learning for vision-language models.
\newblock In \emph{CVPR}, pages 16816--16825, 2022{\natexlab{a}}.

\bibitem[Zhou et~al.(2022{\natexlab{b}})Zhou, Yang, Loy, and Liu]{CoOp}
Kaiyang Zhou, Jingkang Yang, Chen~Change Loy, and Ziwei Liu.
\newblock Learning to prompt for vision-language models.
\newblock \emph{IJCV}, 130\penalty0 (9):\penalty0 2337--2348, 2022{\natexlab{b}}.

\bibitem[Zhou et~al.(2023{\natexlab{a}})Zhou, Gu, Pang, Feng, Cheng, Lu, Shi, and Ma]{SAD}
Qianyu Zhou, Qiqi Gu, Jiangmiao Pang, Zhengyang Feng, Guangliang Cheng, Xuequan Lu, Jianping Shi, and Lizhuang Ma.
\newblock Self-adversarial disentangling for specific domain adaptation.
\newblock \emph{TPAMI}, 45\penalty0 (7):\penalty0 8954--8968, 2023{\natexlab{a}}.

\bibitem[Zhou et~al.(2023{\natexlab{b}})Zhou, Fan, Luo, and Zhang]{NSA}
Wenzhang Zhou, Heng Fan, Tiejian Luo, and Libo Zhang.
\newblock Unsupervised domain adaptive detection with network stability analysis.
\newblock In \emph{ICCV}, pages 6986--6995, 2023{\natexlab{b}}.

\bibitem[Zhu et~al.(2023)Zhu, Guo, Wei, Zhao, and Wu]{D2-UDA}
Yangguang Zhu, Ping Guo, Haoran Wei, Xin Zhao, and Xiangbin Wu.
\newblock Disentangled discriminator for unsupervised domain adaptation on object detection.
\newblock In \emph{IROS}, pages 5685--5691, 2023.

\bibitem[Zhu et~al.(2021)Zhu, Feng, Zhao, Wang, and Li]{Parallel-adapter}
Yaoming Zhu, Jiangtao Feng, Chengqi Zhao, Mingxuan Wang, and Lei Li.
\newblock Serial or parallel? plug-able adapter for multilingual machine translation.
\newblock \emph{arXiv preprint arXiv:2104.08154}, 2021.

\end{thebibliography}
}

\newpage
\section{Appendix}

\subsection{Datasets}
\label{Dataset}
\noindent\textbf{Cross-Weather} Cityscapes~\cite{Cityscapes} contains diverse street scenes captured by a mobile camera in daylight.
The regular partition consists of 2,975 training and 500 validation images annotated with eight classes. 
Foggy Cityscapes~\cite{FoggyCityscapes} simulates three distinct densities of fog on Cityscapes, containing 8,925 training images and 1,500 validation images. 
A standard configuration for cross-weather adaptation is to take the training set of Cityscapes as the source domain and the training set of foggy Cityscapes as the target domain, evaluating cross-weather adaptation performance on the 1500-sized validation set in all eight categories.

\noindent\textbf{Cross-FoV} KITTI~\cite{KITTI} is a crucial dataset for self-driving that includes 7,481 photos annotated with cars.
Collected by driving in rural areas and on highways, it provides data with a different Field of View (FoV). 
To fairly compare with other methods, we migrate KITTI to Cityscapes solely on the car category.

\noindent\textbf{Sim-to-Real} The synthetic dataset SIM10k~\cite{SIM10K} has 10,000 photos from the Grand Theft Auto V video game with labelled bounding boxes in the class car.
We follow existing works to perform this sim-to-real adaptation and report the performance on the class car.

\noindent\textbf{Cross-Style} Pascal VOC~\cite{pascal} is a widely-used real-world dataset containing 2007 and 2012 subsets, annotated with 20 classes. 
Clipart~\cite{clipart} is collected from the website with 1000 comical images, providing bounding box annotations with same classes as Pascal VOC. 
Following the mainstream splitting, we use Pascal VOC 2007 and 2012 train-val split with a total of 16,551 images as the source domain and all Clipart images as the target domain. 

\subsection{Implementation Details}
\label{Implementation}
Following~\cite{DA-Pro}, we adapt RegionCLIP(ResNet-50~\cite{Resnet}) with a domain classifier~\cite{DA-Faster} as the baseline.
We use the Faster-RCNN~\cite{FasterRCNN} as the detector with the default configurations.
Following~\cite{DA-Faster}, one batch of source images with ground truth and one batch of target domain images are forwarded to the proposed DA-Ada in each iteration to calculate the detection, adversarial and decoupling loss.
The hyperparameter $\lambda_{dia}, \lambda_{dita}, \lambda_{dec}$ is set to $0.1$, $1.0$ and $0.1$, respectively.
We set the batch size of each domain to 8 and use the SGD optimizer with a warm-up learning rate.
Mean Average Precision (mAP) with a threshold of 0.5 is taken as the evaluation metric. 
All experiments are deployed on 8 Tesla V100 GPUs. 

\subsection{Comparison to SOTA methods}
\label{SOTA}
\begin{table*}[t]
\centering
\caption{Comparison ($\%$) with existing methods on Cross-Weather adaptation Cityscapes$\rightarrow$Foggy Cityscapes (C$\rightarrow$F), Cross-Fov adaptation KITTI$\rightarrow$Cityscapes (K$\rightarrow$C) and Sim-to-Real adaptation SIM10K$\rightarrow$Cityscapes (S$\rightarrow$C).}
\label{tab-appendix-1}
\setlength\tabcolsep{6pt}
\resizebox{0.95\textwidth}{!}{ 
\begin{tabular}{cccccccccccc}
\toprule
   &\multicolumn{9}{c}{C$\rightarrow$F} & K$\rightarrow$C & S$\rightarrow$ C  \\
    \cmidrule(r){2-10} \cmidrule(r){11-11} \cmidrule(r){12-12}
Methods&Person&Rider&Car&Truck&Bus&Train&Motor&Bicycle&mAP&mAP&mAP\\
\midrule
DA-Faster~\cite{DA-Faster} &29.2 &40.4 &43.4 &19.7 &38.3 &28.5 &23.7 &32.7 &32.0 &41.9 & 38.2\\
VDD~\cite{VD}       &33.4 &44.0 &51.7 &33.9 &52.0 &34.7 &34.2 &36.8 &40.0&-&-\\
DSS~\cite{DSS}       &42.9 &51.2 &53.6 &33.6 &49.2 &18.9 &36.2 &41.8 &40.9 &42.7 & 44.5\\
MeGA~\cite{MEGA-CDA}      &37.7 &49.0 &52.4 &25.4 &49.2 &46.9 &34.5 &39.0 &41.8&43.0 & 44.8\\
SCAN~\cite{SCAN} & 41.7& 43.9& 57.3& 28.7& 48.6& 48.7& 31.0& 37.3& 42.1&45.8&52.6 \\
TIA~\cite{TIA}        &52.1& 38.1& 49.7& 37.7 &34.8 &46.3 &48.6 &31.1 &42.3&44.0 & -\\
DDF~\cite{DDF}        &37.6& 45.5& 56.1& 30.7 &50.4 &47.0 &31.1 &39.8 &42.3&46.0 & 44.3\\
SIGMA~\cite{SIGMA}     &44.0 &43.9 &60.3 &31.6 &50.4 &51.5 &31.7 &40.6 &44.2 &45.8 & 53.7\\
SIGMA++~\cite{sigma++} & 46.4 & 45.1 & 61.0 & 32.1 & 52.2 & 44.6 & 34.8 & 39.9 & 44.5 &49.5 & 57.7\\
CIGAR~\cite{CIGAR} &46.1 &47.3 &62.1 &27.8 &56.6 &44.3 &33.7 &41.3 &44.9 & 48.5& 58.5\\
SAD~\cite{SAD} & 38.3 & 47.2 & 58.8 & 34.9 & 57.7 & 48.3 & 35.7 & 42.0 & 45.2 & - & 49.2 \\
OADA~\cite{OADA} &47.8 &46.5 &62.9 &32.1 &48.5 &50.9 &34.3 &39.8 &45.4& 47.8& 59.2\\
CSDA~\cite{CSDA} &46.6 &46.3 &63.1 &28.1 &56.3 &53.7 &33.1 &39.1 &45.8& 48.6& 57.8\\
HT~\cite{HT} &52.1 &55.8 &67.5 &32.7 &55.9 &49.1 &40.1 &50.3 &50.4 & 60.3 & 65.5\\
D$^2$-UDA~\cite{D2-UDA} &46.9 &53.3 &64.5 &38.9 &61.0 &48.5 &42.6 &54.2 &50.6 & 60.3 & 58.1\\
AT~\cite{AT} &56.3 &51.9 &64.2 &38.5 &45.5 &55.1 &\textbf{54.3} &35.0 &50.9 &-&- \\
NSA-UDA~\cite{NSA}  &50.3 &60.1 &67.7 &37.4 &57.4 &46.9 &47.3 &54.3 &52.7& 55.6& 56.3\\
DA-Pro~\cite{DA-Pro}  &55.4 &62.9 &70.9 &40.3 &63.4 &54.0 &42.3 &58.0 &55.9& 61.4& 62.9\\
\midrule
DA-Ada(Ours)  &\textbf{57.8} &\textbf{65.1} &\textbf{71.3} &\textbf{43.1} &\textbf{64.0} &\textbf{58.6} &48.8 &\textbf{58.7} &\textbf{58.5}($\pm$0.2)& \textbf{66.7}($\pm$0.3)& \textbf{67.3}($\pm$0.2)\\
\bottomrule
\end{tabular}  
}
\end{table*}

\begin{table*}[t]
\centering
\caption{Comparison ($\%$) with existing methods on Cross-Style adaptation task Pascal VOC$\rightarrow$Clipart}
\label{tab-appendix-2}
\setlength\tabcolsep{1.5pt}
\resizebox{0.95\textwidth}{!}{      
\begin{tabular}{cccccccccccccccccccccc}
\toprule
Methods& \rotatebox{90}{Aero} & \rotatebox{90}{Bike}&\rotatebox{90}{Bird}&\rotatebox{90}{Boat}&\rotatebox{90}{Bottle}&\rotatebox{90}{Bus}&\rotatebox{90}{Car}&\rotatebox{90}{Cat}&\rotatebox{90}{Chair}&\rotatebox{90}{Cow}&\rotatebox{90}{Table}&\rotatebox{90}{Dog}&\rotatebox{90}{Horse}&\rotatebox{90}{Motor}&\rotatebox{90}{Person}&\rotatebox{90}{Plant}&\rotatebox{90}{Sheep}&\rotatebox{90}{Sofa}&\rotatebox{90}{Train}&\rotatebox{90}{Tv}&mAP\\ 
\midrule
UaDAN~\cite{UaDAN} & 35.0 & 73.7 & 41.0 & 24.4 & 21.3 & 69.8 & 53.5 & 2.3 & 34.2 & 61.2 & 31.0 & \textbf{29.5} & 47.9 & 63.6 & 62.2 & 61.3 & 13.9 & 7.6 & 48.6 & 23.9 & 40.2 \\
TFD~\cite{TFD} & 27.9 & 64.8 & 28.4 & 29.5 & 25.7 & 64.2 & 47.7 & 13.5 & 47.5 & 50.9 & 50.8 & 21.3 & 33.9 & 60.2 & 65.6 & 42.5 & 15.1 & 40.5 & 45.5 & 48.6 & 41.2 \\
DBGL~\cite{DBGL} & 28.5 & 52.3 & 34.3 & 32.8 & 38.6 & 66.4 & 38.2 & 25.3 & 39.9 & 47.4 & 23.9 & 17.9 & 38.9 & 78.3 & 61.2 & 51.7 & 26.2 & 28.9 & 56.8 & 44.5 & 41.6 \\
IIPD~\cite{IIPD} & 41.5 & 52.7 & 34.5 & 28.1 & 43.7 & 58.5 & 41.8 & 15.3 & 40.1 & 54.4 & 26.7 & 28.5 & 37.7 & 75.4 & 63.7 & 48.7 & 16.5 & 30.8 & 54.5 & 48.7 & 42.1 \\
FGRR~\cite{FGRR} & 30.8 & 52.1 & 35.1 & 32.4 & 42.2 & 62.8 & 42.6 & 21.4 & 42.8 & 58.6 & 33.5 & 20.8 & 37.2 & 81.4 & 66.2 & 50.3 & 21.5 & 29.3 & \textbf{58.2} & 47.0 & 43.3 \\
UMT~\cite{UMT} & 39.6 & 59.1 & 32.4 & 35.0 & 45.1 & 61.9 & 48.4 & 7.5 & 46.0 & \textbf{67.6} & 21.4 & \textbf{29.5} & 48.2 & 75.9 & 70.5 & \textbf{56.7} & 25.9 & 28.9 & 39.4 & 43.6 & 44.1 \\
SIGMA~\cite{SIGMA} & 40.1 & 55.4 & 37.4 & 31.1 & 54.9 & 54.3 & 46.6 & 23.0 & 44.7 & 65.6 & 23.0 & 22.0 & 42.8 & 55.6 & 67.2 & 55.2 & 32.9 & \textbf{40.8} & 45.0 & 58.6 & 44.5 \\
ATMT~\cite{ATMT}  & 37.5 & 63.4 & 37.9 & 29.8 & 45.1 & 62.7 & 41.2 & 19.5 & 43.7 & 57.4 & 22.9 & 25.3 & 39.6 & 87.1 & 70.9 & 50.6 & 29.1 & 32.2 & 58.4 & 50.5 & 45.2 \\
CIGAR~\cite{CIGAR} &35.2 &55.0 &39.2 &30.7 &60.1 &58.1 &46.9 &31.8 &47.0 &61.0 &21.8 &26.7 &44.6 &52.4 &68.5 &54.4 &31.3 &38.8 &56.5 &63.5 &46.2\\
TIA~\cite{TIA} & 42.2 & 66.0 & 36.9 & 37.3 & 43.7 & \textbf{71.8} & 49.7 & 18.2 & 44.9 & 58.9 & 18.2 & 29.1 & 40.7 & \textbf{87.8} & 67.4 & 49.7 & 27.4 & 27.8 & 57.1 & 50.6 & 46.3 \\
SIGMA++~\cite{sigma++} & 36.3 & 54.6 & 40.1 & 31.6 & 58.0 & 60.4 & 46.2 & \textbf{33.6} & 44.4 & 66.2 & 25.7 & 25.3 & 44.4 & 58.8 & 64.8 & 55.4 & 36.2 & 38.6 & 54.1 & \textbf{59.3} & 46.7 \\
CMT~\cite{CMT} & 39.8 & 56.3 & 38.7 & 39.7 & \textbf{60.4} & 35.0 & \textbf{56.0} & 7.1 & \textbf{60.1} & 60.4 & \textbf{35.8} & 28.1 & \textbf{67.8} & 84.5 & \textbf{80.1} & 55.5 & 20.3 & 32.8 & 42.3 & 38.2 & 47.0 \\
\midrule
DA-Ada(Ours) & \textbf{42.3} & \textbf{75.1} & \textbf{48.9} & \textbf{45.9} & 49.0 & \textbf{71.8} & 55.6 & 15.4 & 50.7 & 56.6 & 19.9 & 20.6 & 61.3 & 80.7 & 73.0 & 29.2 & \textbf{37.5} & 21.5 & 52.5 & 52.9 & \textbf{48.0}($\pm$0.1) \\
\bottomrule
\end{tabular}  
}
\end{table*}

We present representative state-of-the-art DAOD approaches for comparison, including feature alignment and semi-supervised learning methods.

\noindent\textbf{Cross-Weather Adaptation Scenario}
Table~\ref{tab-appendix-1} (C$\rightarrow$F) illustrates that the proposed DA-Ada surpasses SOTA DA-Pro~\cite{DA-Pro} by a remarkable margin of $2.6\%$, achieving the highest mAP over eight classes of $58.5\%$. 
Compared with existing methods, our method significantly improves seven categories (\ie ~person, rider, car, truck, bus, train, and bicycle) ranging from $0.4\%$ to $5.3\%$.
Compared with the SOTA decoupling method D$^2$-UDA~\cite{D2-UDA}, the DA-Ada attains an improvement of $7.9\%$ and promotes $3.0\sim 10.9\%$ on all categories.
The superior performance shows modifying domain-invariant knowledge with domain-specific knowledge could enhance the discriminative capability on the target domain.

\noindent\textbf{Cross-FOV Adaptation Scenario}
Table~\ref{tab-appendix-1} (K$\rightarrow$C) indicates a noticeable $5.3\%$ improvement on SOTA DA-Pro~\cite{DA-Pro} by the DA-Ada, reaching an astounding peak of $66.7\%$ mAP.
As K$\rightarrow$C adaptation faces more complicated shape confusion than C$\rightarrow$F, it requests higher discriminability of the model. 
Therefore, the considerable enhancement validates that the proposed method can efficiently improve the discriminability of the visual encoder in new scenarios. 

\noindent\textbf{Sim-to-Real Adaptation Scenario}
We report the experimental results on SIM10k $\rightarrow$ Cityscapes benchmark in Table~\ref{tab-appendix-1} (S$\rightarrow$C).
The proposed DA-Ada achieves the best results of $67.3\%$ mAP, outperforming the previous best entry HT~\cite{HT} $65.5\%$ with $1.8\%$.
The performance of DA-Ada is superior in the difficult adaptation task, which further demonstrates that our strategy is robust not only in appearance but also in more complex semantics adaptation tasks.

\noindent\textbf{Cross-Style Adaptation Scenario}
Additionally, we assess DA-Ada on the more challenging Cross-Style adaptation, where the semantic hierarchy has a broader domain gap. 
DA-Ada peaks with $48.0\%$, outperforming all the SOTA methods presented in Table~\ref{tab-appendix-2}.
It demonstrates that injecting cross-domain information into the visual encoder could benefit the adaptation.
Especially, DA-Ada exceeds all the compared methods on six categories (aeroplane, bike, bird, boat, bus, and sheep), which verifies the method is effective under challenging domain shifts and in multi-class problem scenarios.

\subsection{Sensitivity on $\mathcal{L}_{dia}$}
\begin{table}[h]
\centering
\caption{ Sensitivity to hyper-parameters of initialization of $\lambda_{dia}$.}
\label{L_dia}
\setlength\tabcolsep{10pt}
\begin{tabular}{ccccccc}
\toprule
\multicolumn{7}{c}{Cityscapes$\rightarrow$ FoggyCityscapes} \\
\midrule
$\lambda_{dia}$ & 0.01 & 0.05 & 0.1 & 0.5 & 1.0 & 10.0\\
\midrule
mAP  &  57.1  & 57.8 & 58.5 & 58.1 & 58.0 & 53.4 \\ 
\bottomrule
\end{tabular}  
\end{table}
To select hyper-parameters for the adversarial loss in DIA, we perform experiments of different choices of the weight value $\lambda_{dia}$.
We conduct the experiment on DA-Ada on Cityscapes$\rightarrow$FoggyCityscapes adaptation scenarios, as shown in Table~\ref{L_dia}.
Initialized with $0.01$, the DIA suffers from insufficient learning of domain-invariant knowledge, only attaining $57.1\%$ mAP.
When initialized with $0.05\sim 1.0$, the performance of DA-Ada is similar and achieves the best of $58.5\%$ with $\lambda_{dia}=1.0$.
Increasing the $\lambda_{dia}$ to $10.0$ sufferers significant performance degradation.
We attribute this to the model focusing too much on alignment rather than detection.

\subsection{Sensitivity on $\lambda_{dita}$}
\begin{table}[h]
\centering
\caption{ Sensitivity to hyper-parameters of initialization of $\mathcal{L}_{dita}$.}
\label{L_dita}
\setlength\tabcolsep{10pt}
\begin{tabular}{cccccc}
\toprule
\multicolumn{6}{c}{Cityscapes$\rightarrow$ FoggyCityscapes} \\
\midrule
$\lambda_{dita}$ & 0.1 & 0.5 & 1.0 & 5.0 & 10.0 \\
\midrule
mAP  &  58.1  & 58.2 & 58.5 & 57.9 & 56.9 \\ 
\bottomrule
\end{tabular}  
\end{table}
We also explicitly study the sensitivity of weight value $\lambda_{dita}$ for the adversarial loss in visual-guided domain prompt, as shown in Table~\ref{L_dita}.
As the weight value increases, the performance peaks with $\lambda_{dita}=1.0$ and then appears to decline.
Since the hand-crafted token "A photo of [CLASS]" provides a solid prior, the learnable prompt is better initialized in the early stages of training.
Therefore, the proposed model is more robust to the $\lambda_{dita}$ compared to $\lambda_{dia}$.

\subsection{Sensitivity on $\lambda_{dec}$}
\begin{table}[h]
\centering
\caption{Sensitivity to hyper-parameters of initialization of $\mathcal{L}_{dec}$.}
\label{L_dec}
\setlength\tabcolsep{10pt}
\begin{tabular}{cccccc}
\toprule
\multicolumn{6}{c}{Cityscapes$\rightarrow$ FoggyCityscapes} \\
\midrule
$\lambda_{dec}$ & 0.01 & 0.05 & 0.1 & 0.5 & 1.0 \\
\midrule
mAP  &  57.5  & 57.9 & 58.5 & 58.3 & 58.4 \\ 
\bottomrule
\end{tabular}  
\end{table}
We also evaluate the sensitivity of weight value $\lambda_{dec}$ for the decouple loss between DIA and DSA, as shown in Table~\ref{L_dec}.
As the weight value increases, the performance rises rapidly until $\lambda_{dec}=1.0$ and then declines smoothly.
$\mathcal{L}_{dec}$ decouples domain-invariant and domain-specific knowledge by driving DIA to be orthogonal to the features extracted by DSA.
Therefore, applying the decoupling loss with the same scale as the adversarial loss can optimize the goal relatively stably.

\subsection{Ablation for Multi-scale Down-projector $\mathcal{C}^D$}
\begin{table}[h]
\centering
\caption{ Ablation for Different Resolution in Multi-scale Down-projector $\mathcal{C}^D$.}
\label{tab-scale}
\setlength\tabcolsep{22pt}
\resizebox{0.8\textwidth}{!}{      
\begin{tabular}{cccccc}
\toprule
\multicolumn{6}{c}{Cityscapes$\rightarrow$ FoggyCityscapes} \\
\midrule \multicolumn{5}{c}{Resolutions} & \multirow{3}*{mAP} \\
\cmidrule(r){1-5}
$1$ & $\frac{1}{2}$ & $\frac{1}{4}$ & $\frac{1}{8}$ & $\frac{1}{16}$ & \\
\midrule
$\checkmark$ & & & & & 57.6\\
$\checkmark$ & $\checkmark$ & & & & 57.8\\
$\checkmark$ & $\checkmark$ & $\checkmark$ & & & 58.3\\
$\checkmark$ & $\checkmark$ & $\checkmark$ & $\checkmark$ & & 58.5\\
$\checkmark$ & $\checkmark$ & $\checkmark$ & $\checkmark$ & $\checkmark$ & 58.2\\
\bottomrule
\end{tabular}  }
\end{table}
Table~\ref{tab-scale} shows the impact of the different resolutions in multi-scale down-projector $\mathcal{C}^D$.
The results indicate that while introducing various resolutions contributes to modeling multi-scale domain knowledge, inappropriate receptive fields may harm feature extraction performance.
Concretely, we observed that the mAP on the target domain (Foggy Cityscapes) peaks when the number of down-projectors $M$ is set to $4$, and the scaling ratios are $\{1, \frac12, \frac14, \frac18\}$, respectively.
And further applying $\frac{1}{16}$ to $\mathcal{C}^D$ results in slightly performance degradation.
These experiments suggest that applying a single convolution or introducing excessive distinct resolutions is unsuitable for learning domain knowledge, and the choice of resolution requires consideration of the difference in scale between the source and target domains.

\subsection{Image or Instance-level Visual-guided Textual Adapter?}
We explore whether to apply the visual-guided textual adapter at the image or instance levels.
In DITA and DSTA, we replace the proposal embedding with the entire image as visual input, and it achieves $57.8\%$, suffering a performance drop of $0.7\%$.
This indicates that instance-level alignment avoids the influence of background on learning domain knowledge in the foreground.

\subsection{Evaluation on multiple Baselines}
\begin{table}[h]
\setlength\tabcolsep{2pt}
\centering
\caption{ Results of multiple Baselines on C$\rightarrow$F adaptation.}
\label{tab-baseline}
\begin{tabular}{cccc}
\toprule
Baseline & w/o DA-Ada & with DA-Ada & Gains \\ 
\midrule
DSS     &40.9 & 48.1 &  +7.2\\ 
CLIP+Faster-RCNN & 42.8& 52.6& +9.8\\
\bottomrule
\end{tabular}  
\end{table}
To properly evaluate the method, we introduce Da-Ada to two weaker baselines in~\ref{tab-baseline}.
We inject DA-Ada into DSS and attain $7.2\%$ improvement on mAP, showing great efficiency in feature-alignment methods.
For further validation, we first adapt the classification model CLIP to Faster-RCNN to build a vanilla VLM detector with $42.8\%$ mAP.
Then we freeze the backbone and attach DA-Ada to the detector, achieving $52.6\%$ mAP with an improvement of $9.8\%$.
Experiments show that even with weak baselines, the proposed DA-Ada shows competitiveness to SOTA methods.

\subsection{Performance and Computational Overhead}
\label{overhead}
\begin{table}[h]
\setlength\tabcolsep{2pt}
\centering
\caption{Comparison on performance and computational overhead on C$\rightarrow$F adaptation.}
\label{tab-appendix-3}
\resizebox{0.95\textwidth}{!}{      
\begin{tabular}{ccccccc}
\hline
Method & mAP & Inference time(s)/iter  & Training time(s)/iter & Total iter & Mem usg.(MB) \\ 
\hline
Global Fine-tune & 53.6 & 0.40 & 2.67  & 25000 &  12977 \\ 
DA-Pro           & 54.6 & 0.40 & 1.47  & 1000  &  4034 \\
DA-Ada           & 58.5 & 0.42 & 1.61  & 2500  &  6046 \\
\hline
\end{tabular}  }
\end{table}
To verify the effectiveness of DA-Ada, we compare the performance and computational overhead with global fine-tune and DA-Pro on C$\rightarrow$F adaptation.
We initial the three methods with the same VLM backbone.
Global fine-tuning has the largest time and memory overhead but only achieves the lowest performance, indicating the limitations of traditional DAOD methods in optimizing VLM.
Compared with global fine-tuning, DA-Pro significantly reduces time and memory overhead while improving performance.
Furthermore, DA-Ada significantly improves mAP with $4.9\%$ while only using $6\%$ of the time and $47\%$ of the memory, showing great efficiency in adapting cross-domain information to VLM.

\subsection{Failure Cases}
\begin{figure*}[h!]
  \centering
  \includegraphics[width=1.0\textwidth]{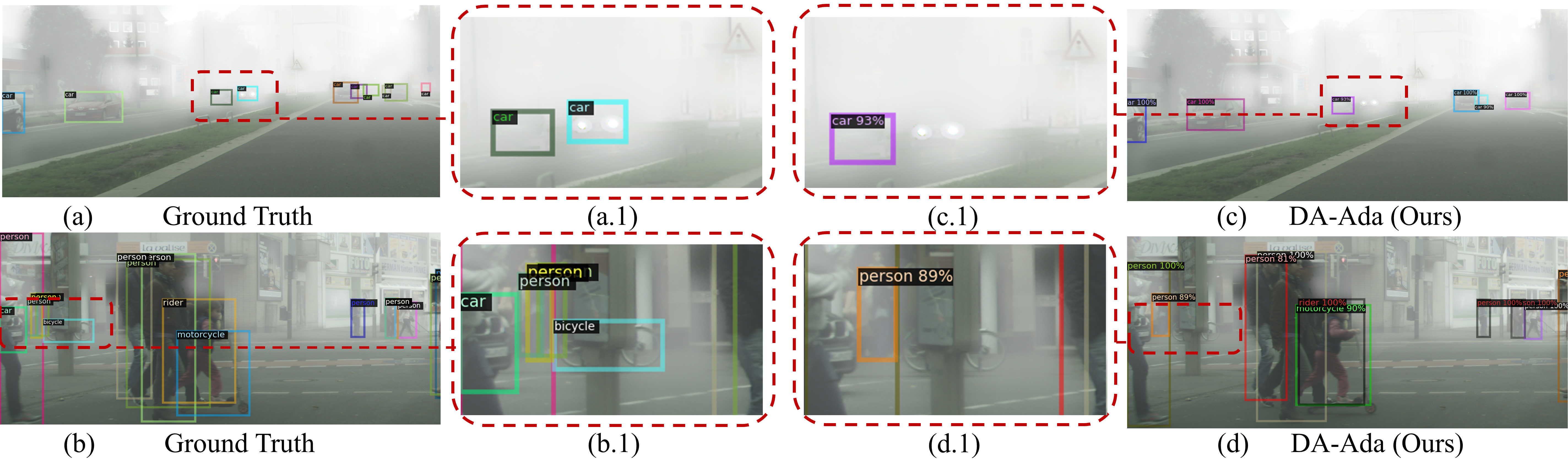}
  \vspace{-2pt}
  \caption{Examples of failure cases on the Cross-Weather adaptation scenario. We visualize the ground truth (a)(b) and the detection boxes of DA-Ada (c)(d). (a.1)(b.1)(c.1)(d.1) are zoomed from corresponding region for better view.}
  \label{fig5}
  \vspace{-10pt}
\end{figure*}

We provide some examples of failure cases on the Cross-Weather adaptation scenario in Fig~\ref{fig5}. 
We visualize the ground truth (a)(b) and the detection boxes of DA-Ada (c)(d). 
In (c.1), DA-Ada misses the car with its headlights on in the fog. Since the source data Cityscapes is collected on sunny days, few cars turned on their lights in the training set. 
Therefore, DA-Ada missed such out-of-distribution data. In (d.1), DA-Ada misses the bicycle and person blocked by other foreground objects. 
Since occlusion causes great damage to semantics, this type of missed detection is widely seen in object detection methods.

\subsection{Limitation}
\label{limitation}
Though effective, the proposed DA-Ada is specially designed for the domain adaptive object detection task, where a labelled source domain and a unlabelled target domain are needed.
Currently, the method cannot deal with the setting of multiple source domains or no target domain.
We plan to resolve these problems in our future research.

\newpage

\end{document}